\documentclass[11pt]{article}

\usepackage[final]{acl}

\usepackage{times}
\usepackage{latexsym}

\usepackage[T1]{fontenc}

\usepackage[utf8]{inputenc}

\usepackage{microtype}

\usepackage{inconsolata}
\usepackage{enumitem}

\usepackage{graphicx}
\usepackage{pifont}
\usepackage{amssymb}
\usepackage{booktabs}

\usepackage{array}
\usepackage{ragged2e}
\usepackage{longtable}
\usepackage{ragged2e, supertabular}
\usepackage[table]{xcolor}
\usepackage{multirow}

\usepackage{xcolor}
\definecolor{subtitlegray}{RGB}{90, 90, 90}
\usepackage{graphicx,pifont}

\usepackage{stfloats}
\usepackage[table]{xcolor}
\usepackage{tabularx}
\usepackage{hyperref}

\newcommand{\ours}{TimeSage-EV}

\newcommand{\logo}[1]{\raisebox{-0.25ex}{\includegraphics[height=0.95em]{#1}}\,}
\newcommand{\val}[3]{\ensuremath{#1^{#2}_{#3}}}
\newcommand{\valb}[3]{\ensuremath{\mathbf{#1}^{\mathbf{#2}}_{\mathbf{#3}}}}

\title{\ours: A Live Benchmark for Agentic Time Series Analysis in \underline{Ev}olving Environments
}

\author{
\textbf{Qingren Yao}\textsuperscript{1*},\
\textbf{Yaxuan Kong}\textsuperscript{2,3*},\
\textbf{Yuqi Nie},\
\textbf{Yichen Li}\textsuperscript{3},\
\textbf{Stefan Zohren}\textsuperscript{2},\
\textbf{Anna Vettoruzzo}\textsuperscript{1}, \\
\textbf{Qingsong Wen}\textsuperscript{4},\
\textbf{Ming Jin}\textsuperscript{5\dag},\
\textbf{Joaquin Vanschoren}\textsuperscript{1\dag} \\
\textsuperscript{1}Eindhoven University of Technology,\
\textsuperscript{2}University of Oxford,\
\textsuperscript{3}VulpiVox Intelligence, \\
\textsuperscript{4}Squirrel Ai Learning,\
\textsuperscript{5}Griffith University \\
q.yao@tue.nl; yaxuan.kong@eng.ox.ac.uk; mingjinedu@gmail.com; j.vanschoren@tue.nl\\[0.4em]
\href{https://huggingface.co/datasets/TimeSage-Series/TimeSage-EV}{\logo{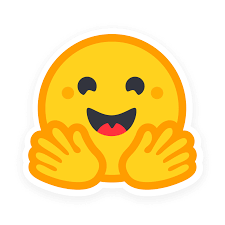}\,Dataset}
\quad
\href{https://github.com/TimeSage-Series/TimeSage-EV}{\logo{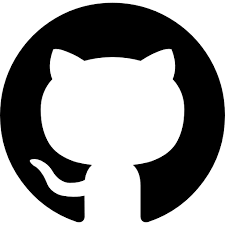}\,Code}
}

\usepackage[most]{tcolorbox}

\newtcblisting{promptbox}[1][]{
  enhanced,
  breakable,
  colback=gray!5!white,      
  colframe=gray!60!black,    
  boxrule=0.5pt,             
  arc=3pt,                   
  left=4pt, right=4pt, top=4pt, bottom=4pt,
  listing only,
  listing options={
    basicstyle=\small\ttfamily,
    breaklines=true,         
    breakatwhitespace=true,
    breakindent=0pt,         
    breakautoindent=false,   
    columns=fullflexible,
    keepspaces=true,
  },
  #1
}

\begin{document}
\maketitle

\begingroup
\renewcommand\thefootnote{}\footnotetext{%
\textsuperscript{*}Equal contribution.\ \textsuperscript{\dag}Corresponding author.}%
\addtocounter{footnote}{0}
\endgroup

\begin{abstract}
Time series analysis in high-stakes domains relies on recurring data releases, where new observations can alter the evidence base and the validity of later conclusions. Existing time series QA benchmarks mostly rely on fixed snapshots, leaving temporal validity and cutoff-aware evidence use unevaluated. We introduce \textbf{TimeSage-EV}, a live benchmark for agentic time series analysis in evolving environments. It tracks 60 real institutional scenarios across 6 domains, comprising 1,485 scenario-period QA pairs from Feb 2023 to May 2026 and spanning monthly, weekly, daily, and irregular release cadences. At each period, large language model (LLM) agents receive time series data and source reports, while the withheld target release provides ground truth. TimeSage-EV evaluates state identification, data summarization, and outlook reasoning. Experiments with frontier LLM agents and \textbf{TimeSage-1.0}, a novel self-evolving agent with a reusable analytical skill library, reveal significant performance gaps across model tiers and recurring failures in temporal validity, exogenous context use, and adaptation. We release TimeSage-EV as a research resource with monthly updates, code, a leaderboard, and failure-mode analyses.
\end{abstract}

\section{Introduction}
\label{section:introduction}
Across high-stakes domains, from finance and public health to transportation, time series analysis is not a static prediction task but a recurring workflow. Analysts must continuously ingest new data releases, update historical context, and synthesize numeric trends with institutional reports. Crucially, each new release shifts the available evidence distribution, requiring strict adherence to timestamp-aware information cutoffs. We formalize this recurring workflow as agentic time series analysis in evolving environments, requiring Large Language Models (LLMs) to maintain temporal validity while reasoning over dynamic multimodal evidence.

Existing benchmarks only partially cover this setting, with two key limitations. \textit{\textbf{(1)}} Time series question answering (QA) benchmarks, such as ChatTime-TSQA~\cite{wang2025chattime}, Time-MQA~\cite{kong2025timemqa}, and TSRBench~\cite{yu2026tsrbench}, test temporal reasoning over time series, but usually construct fixed QA pairs over predefined time series tasks rather than evolving workflows. \textit{\textbf{(2)}} Agent-centric benchmarks evaluate coding~\cite{jimenez2024swebench}, web navigation~\cite{zhou2024webarena, mialon2024gaia}, multi-hop retrieval~\cite{wei2025browsecomp}, and tool use~\cite{yao2025taubench}, but are not designed around recurring time series analysis. They rarely require agents to track time-indexed evidence, compare new observations with historical context, and synthesize time series with institutional reports. Thus, it remains unclear whether LLM agents can perform such analytical work as the evidence changes over time.

Constructing a benchmark for this dynamic setting presents three challenges.\textit{\textbf{(1)}} Each scenario must be drawn from a stable institutional source with a recurring release schedule, ensuring the evaluation reflects real-time monitoring rather than synthetic snapshots. \textit{\textbf{(2)}} Each target period must impose a strict information cutoff. Agents can only access cumulative data and source documents published prior to the target release, while the target release itself is withheld for ground truth construction and evidence verification. \textit{\textbf{(3)}} Evaluation cannot be restricted to closed-form choices. To align with practical applications, it must assess both key analytical tasks and human-readable reports. These requirements motivate a benchmark design with timestamp-aware ground truth, traceable evidence, cutoff-valid inputs, and metrics that explicitly separate factual correctness from report faithfulness.

\begin{figure*}[htbp]
\begin{center}
\includegraphics[width = \linewidth]{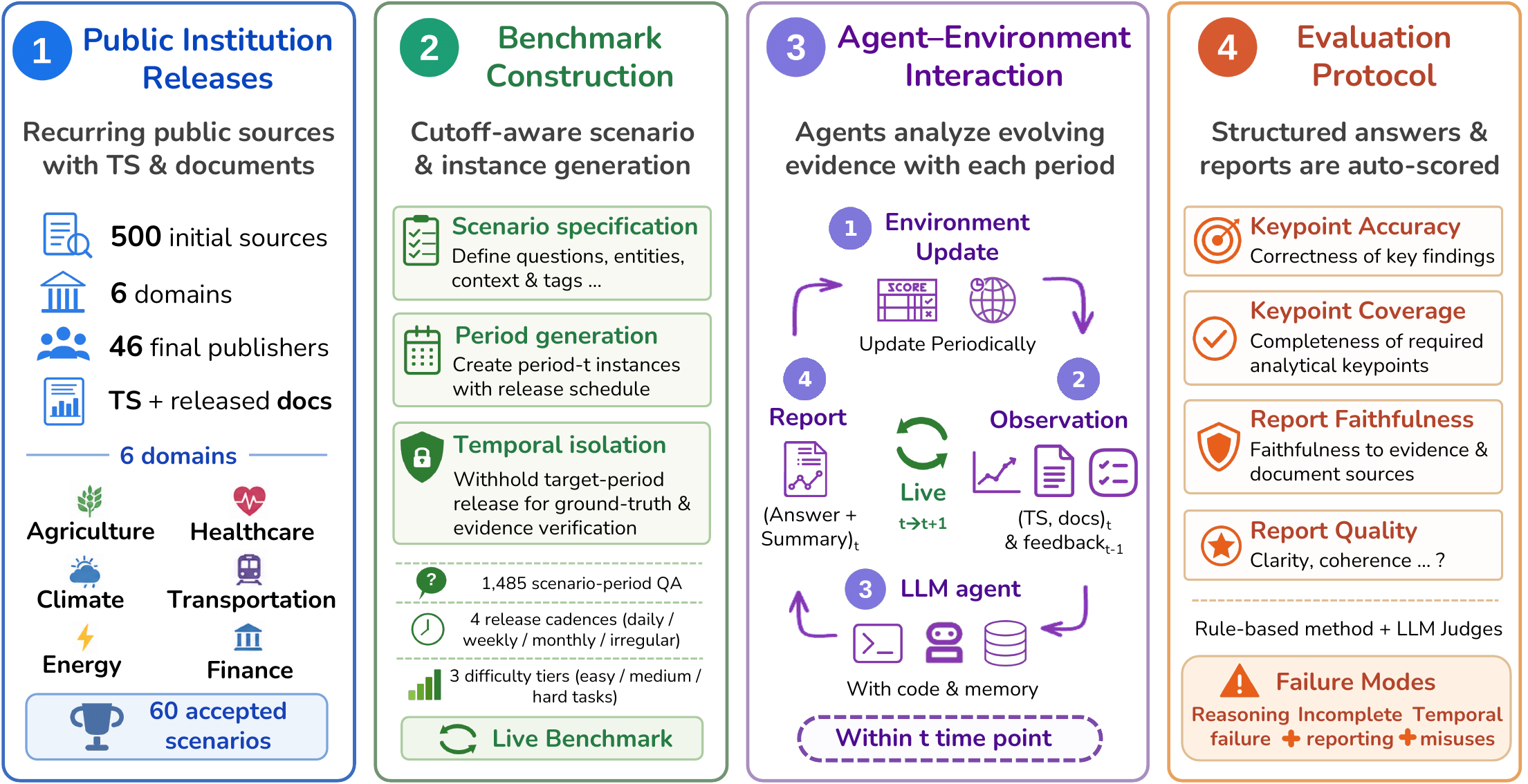}
\end{center}
\vspace{-5pt}
\caption{Overview of TimeSage-EV, supporting agentic time series tasks and a live leaderboard. TS: time series.  }
\label{Figure_1}
\end{figure*}

To address these challenges, we introduce \textbf{TimeSage-EV}, a live benchmark for \textbf{agentic time series analysis in evolving environments}. As shown in Figure~\ref{Figure_1}, each scenario is a recurring task anchored to an institutional source, defined by a fixed domain, release cadence, difficulty tier and set of questions. TimeSage-EV tracks 60 scenarios across six domains from February 2023 to May 2026, covering agriculture, climate, energy, finance, healthcare, and transportation. The scenarios are organized into three difficulty tiers: easy tasks focus on state identification; medium tasks require change detection, composition analysis, and event detection; and hard tasks require attribution and forecast. Each scenario unfolds over monthly, weekly, daily, or irregular release periods.

For each target period, the benchmark instantiates specific questions and structured answer fields, comprising 1,485 scenario-period QA pairs. The agent receives pre-cutoff time series data and documents, while the target-period release is withheld for verification. The question types are annotated with a 27-tag taxonomy grouped into six families (Appendix~\ref{app:sec_data_collection} Table~\ref{tab:question_taxonomy}): state identification, change detection, composition analysis, event detection, attribution, and outlook. The agent answers the questions and compiles them into a report. Critically, as new observations arrive in periods, the agent must update its judgments and revise its report adhering to the rolling evidence cutoff.

Our evaluation protocol scores the output along four axes: keypoint accuracy, keypoint coverage, report faithfulness, and report quality. This distinguishes agents that compute correct answers from those that generate fluent but weakly grounded reports. We then evaluate frontier LLM agents and \textbf{TimeSage-1.0}, a self-evolving agent with a reusable skill library, on \textbf{TimeSage-EV}. The results reveal large performance gaps across difficulty tiers and failures in memory management, temporal misuse, and report synthesis. These findings show that current agents struggle to maintain cutoff-correct evidence and to adapt to new releases, offering diagnostic guidance for reliable agents.

Our contributions are summarized as follows.
\begin{itemize} [leftmargin=1em, itemsep=0.2em, parsep=0em, topsep=0.3em]
    \item We formalize a novel task for \textbf{agentic time series analysis in evolving environments}, requiring temporal isolation, cutoff-valid evidence use, and recurring analytical updates.
    \item We construct \textbf{TimeSage-EV}, a live benchmark\footnote{The benchmark is kept live through a maintenance script that runs automatic updates monthly. See Appendix~\ref{appendix_2}.} comprising 60 scenarios across 6 domains, 3 tiers, and 1,485 scenario-period questions with ground truth and traceable evidence.
    \item We design a rigorous \textbf{multi-axis evaluation protocol} measuring keypoint accuracy, keypoint coverage, report faithfulness, and report quality across structured answers and analytical reports.
    \item We benchmark frontier LLM agents and the diagnostic \textbf{TimeSage-1.0}, uncovering failure modes in memory management, temporal misuse, and report synthesis to guide future research.
\end{itemize}

\section{Related Work}
\label{section:related_work}

Prior work spans language-grounded time series analysis and long-horizon live agent benchmarks. However, live benchmarks for agentic time series analysis in evolving environments remain underexplored, particularly in scenarios where agents must repeatedly analyze recurring data releases while adhering to cutoff-valid numerical and documentary evidence. Table~\ref{tab:dataset} summarizes representative benchmarks across these dimensions.


\newcommand{\yes}{\textcolor{red}{\ding{52}}}
\newcommand{\no}{\ding{56}}
\newcommand{\partialmark}{\(\triangle\)}

\begin{table}[htbp]
\caption{Comparison of representative benchmarks. \yes{} denotes support, \partialmark{} partial support, and \no{} not covered.}
\resizebox{1\linewidth}{!}{
\fontsize{11}{14}\selectfont
\renewcommand{\arraystretch}{1}{
\setlength{\heavyrulewidth}{1.25pt}
\begin{tabular}{lcccccc}
\toprule
\multicolumn{1}{l}{\textbf{Benchmark}} 
& \multicolumn{1}{c}{\textbf{TS}} 
& \multicolumn{1}{c}{\textbf{Text}} 
& \multicolumn{1}{c}{\textbf{Agent}} 
& \multicolumn{1}{c}{\textbf{Live}} 
& \multicolumn{1}{c}{\textbf{Evol.}} 
& \multicolumn{1}{c}{\textbf{Cutoff}} \\ 
\midrule
Time-LLM~\cite{jin2024timellm} & \yes & \no & \no & \no & \no & \no \\
ChatTime~\cite{wang2025chattime} & \yes & \yes & \no & \no & \no & \no \\
Time-MQA~\cite{kong2025timemqa} & \yes & \yes & \no & \no & \no & \no \\
TimeSeriesExam~\cite{cai2024timeseriesexam} & \yes & \yes & \no & \no & \no & \no \\
WebArena~\cite{zhou2024webarena} & \no & \yes & \yes & \no & \no & \no \\
LiveBench~\cite{white2025livebench} & \no & \partialmark & \partialmark & \yes & \no & \no \\
ForecastBench~\cite{karger2025forecastbench} & \no & \no & \partialmark & \yes & \no & \no \\
TimeSeriesGym~\cite{cai2025timeseriesgym} & \yes & \yes & \yes & \no & \no & \no \\
TimeSage-MT~\cite{kong2026timesage} & \yes & \yes & \yes & \no & \no & \no \\
\textbf{TimeSage-EV} & \yes & \yes & \yes & \yes & \yes & \yes \\
\bottomrule
\end{tabular}}}
\vspace{1mm}

\footnotesize{\textit{TS}: numerical time series evidence; \textit{Text}: textual or documentary evidence; \textit{Evol.}: the same scenario is revisited across release periods; \textit{Cutoff}: period-specific evidence isolation.}
\label{tab:dataset}
\vspace{-2mm}
\end{table}

\noindent\textbf{Language-grounded time series analysis.}
Time series analysis is increasingly framed as a language-facing task that connects numerical patterns with instructions and domain context. LLMTime~\cite{gruver2023llmtime} and Time-LLM~\cite{jin2024timellm} recast time series for language-model use, while ChatTime~\cite{wang2025chattime}, ChatTS~\cite{xie2025chatts}, Time-MQA~\cite{kong2025timemqa}, TSRBench~\cite{yu2026tsrbench}, TimeSeriesExam~\cite{cai2024timeseriesexam}, TimeSage-MT~\cite{kong2026timesage}, and TemporalBench~\cite{weng2026temporalbench} evaluate temporal reasoning through natural-language questions. Beyond question answering, TimeSeriesGym~\cite{cai2025timeseriesgym} evaluates LLMs on end-to-end time series machine learning tasks. These settings remain snapshot-based, with fixed questions over static evidence, and do not evaluate how agents update analyses across recurring releases.

\noindent\textbf{Long-horizon live agent benchmarks.}
LLM agents are evaluated through tool use, retrieval, code execution, and memory, as in ReAct~\cite{yao2023react}, Toolformer~\cite{schick2023toolformer}, and AutoGen~\cite{wu2024autogen}. Benchmark tasks span web navigation, dialogue, and software engineering, including WebArena~\cite{zhou2024webarena}, GAIA~\cite{mialon2024gaia}, Tau-Bench~\cite{yao2025taubench}, and SWE-bench~\cite{jimenez2024swebench}. Live benchmarks such as LiveBench~\cite{white2025livebench} and LiveCodeBench~\cite{jain2024livecodebench} reduce contamination through refreshed task pools, while ForecastBench~\cite{karger2025forecastbench} evaluates predictions after outcomes resolve. Yet, these settings remain refreshed snapshots, rather than evolving scenarios with recurring releases and cumulative numerical-documentary evidence.

\noindent\textbf{Positioning TimeSage-EV.}
Existing work leaves a critical gap at the intersection of these two lines: live, continuous agent evaluation for recurring time series analysis. \textbf{TimeSage-EV} addresses this gap by evaluating agentic time series analysis in evolving environments, where agents revisit institutional releases, combine evolving numerical and textual evidence, respect time cutoffs, and produce both structured answers and grounded reports. TimeSage-EV moves beyond static QA task to evaluate evolving time series analysis as a dynamic, long-horizon agentic workflow (Appendix~\ref{appendix_1}).

\section{Methodology}

\subsection{Task Formulation}
\label{sec:task-formulation}

We formulate \textbf{agentic time series analysis in evolving environments} as a recurring task over sequential release periods. In \textbf{agentic time series analysis}, an agent must utilize tools to inspect time series, consult cutoff-valid documents, answer structured questions, and generate an evidence-based report. An \textbf{evolving environment} presents this task under changing information states, as new evidence arrives across subsequent periods (Figure~\ref{Figure_2}). We formalize this setting below and provide expanded definitions in Appendix~\ref{appendix_2}.

\begin{figure}[htbp]
\begin{center}
\includegraphics[width = 1.02\linewidth]{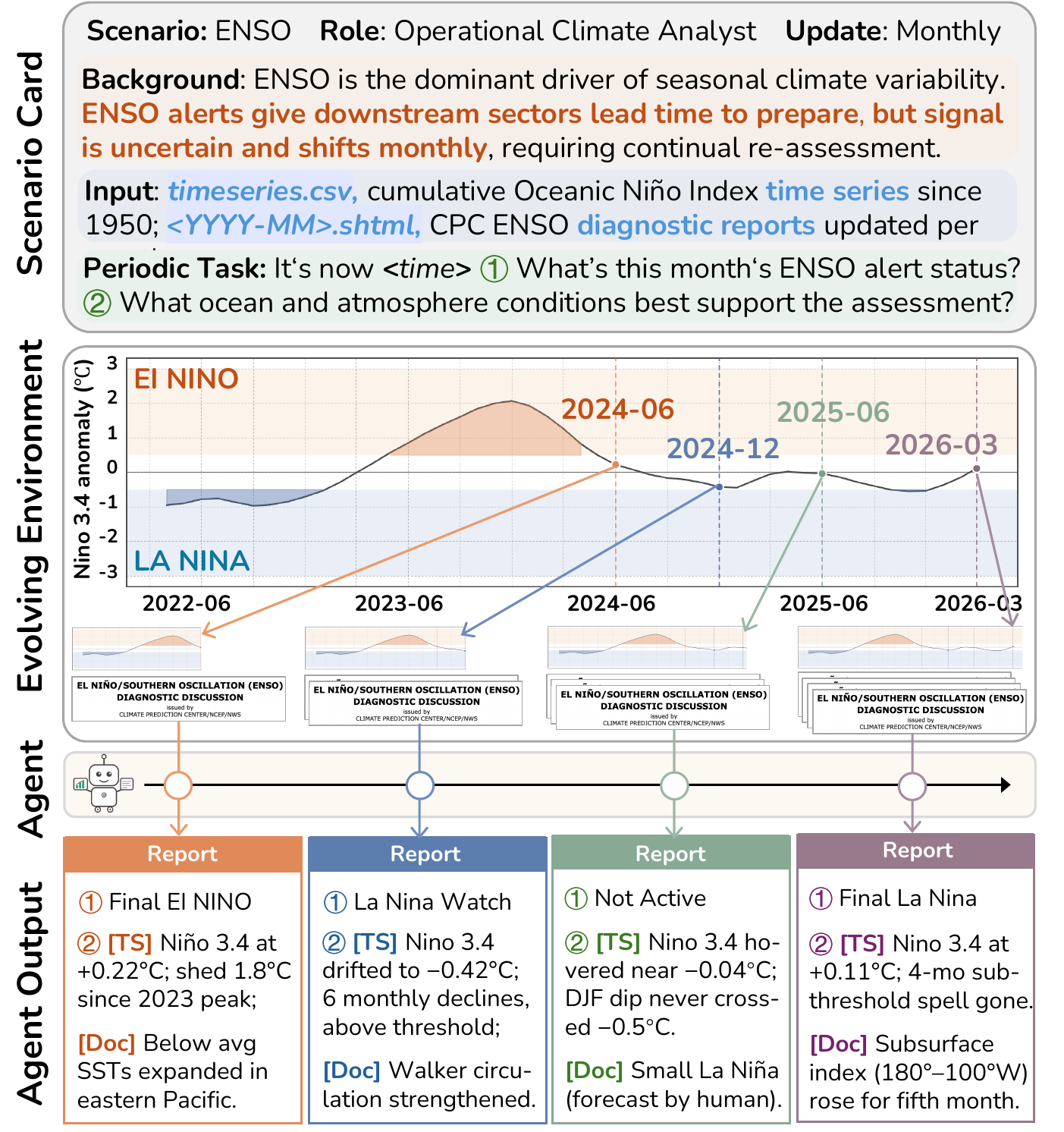}
\end{center}
\vspace{-10pt}
\caption{Evolving environment in TimeSage-EV with ENSO scenario. TS: time series, Doc: source document.}
\vspace{-10pt}
\label{Figure_2}
\end{figure}

\noindent\textbf{Scenario Task.}
An \textbf{environment} for scenario \(s\) defines a recurring analytical task over time:
$$
\mathcal{E}_s = (D_s,\, Q_s,\, A_s,\, \{e_{t_k}^{(s)}\}_{t_k\in\mathcal{T}_s}),
$$
where \(D_s\) denotes the domain background, \(Q_s\) is the recurring question template, \(A_s\) is the structured answer schema, and \(\{e_{t_k}^{(s)}\}\) is the sequence of evidence states across release periods \(\mathcal{T}_s=\{t_1,\ldots,t_T\}\). The task specification \((D_s,Q_s,A_s)\) remains fixed within a scenario, while \(e_{t_k}^{(s)}\) changes as new releases become available.

\noindent\textbf{Period Instance.} For each target period \(t_k\), the period instance is defined by the environment state
$$
e_{t_k}^{(s)}
=
\left(q_{t_k},\, X_{\leq t_k},\, R_{<t_k},\, W_{\leq t_k}^{(s)}\right),
$$
where \(q_{t_k}\) is the natural-language question, \(X_{\leq t_k}\) is the cumulative time series table up through the target period, and \(R_{<t_k}\) contains source documents released strictly before the target period. The target period release \(R_{t_k}\) is withheld from the agent and used only for ground truth construction and evidence verification. Thus, the time series is visible, but the document is not. Finally, \(W_{\leq t_k}^{(s)}\) denotes external web evidence available prior to the cutoff. By default, \(W_{\leq t_k}^{(s)}=\emptyset\), and it is enabled only for scenarios with cutoff-valid retrieval.

\noindent\textbf{Agent State and Output.}
Within this environment, the LLM agent maintains a state \(S_{t_k}\), which may encompass memory, reasoning traces, or reusable analysis artifacts. Given a period instance, the agent returns
$$
\left(\hat{y}_{t_k},\, \hat{r}_{t_k},\, S_{t_{k+1}}\right)
=
\mathrm{Agent}_{s}\!\left(e_{t_k}^{(s)},\, S_{t_k}\right),
$$
where \(\hat{y}_{t_k}\) is a schema-conformant structured answer and \(\hat{r}_{t_k}\) is a cutoff-grounded report. Stateless agents use \(S_{t_k}=\emptyset\). Figure~\ref{Figure_2} illustrates this \textbf{evolving} process. As the benchmark progresses from \(t_k\) to \(t_{k+1}\), the task specification remains fixed, but the evidence state expands:
$$
X_{\leq t_k} \subseteq X_{\leq t_{k+1}},
\qquad
R_{<t_k} \cup \{R_{t_k}\} \subseteq R_{<t_{k+1}}.
$$
Thus, each step introduces a new evidence state in which the time series is extended, the previous target release becomes prior evidence, and the next target release is withheld for evaluation. The agent must continuously revise its judgments without violating the time cutoff. As future releases are published, the same process creates new temporally isolated evaluation periods, making TimeSage-EV a continuously updated \textbf{live benchmark}.

\subsection{Benchmark Construction}
\label{sec:benchmark_construction}
\begin{figure}[htbp]
\begin{center}
\includegraphics[width = \linewidth]{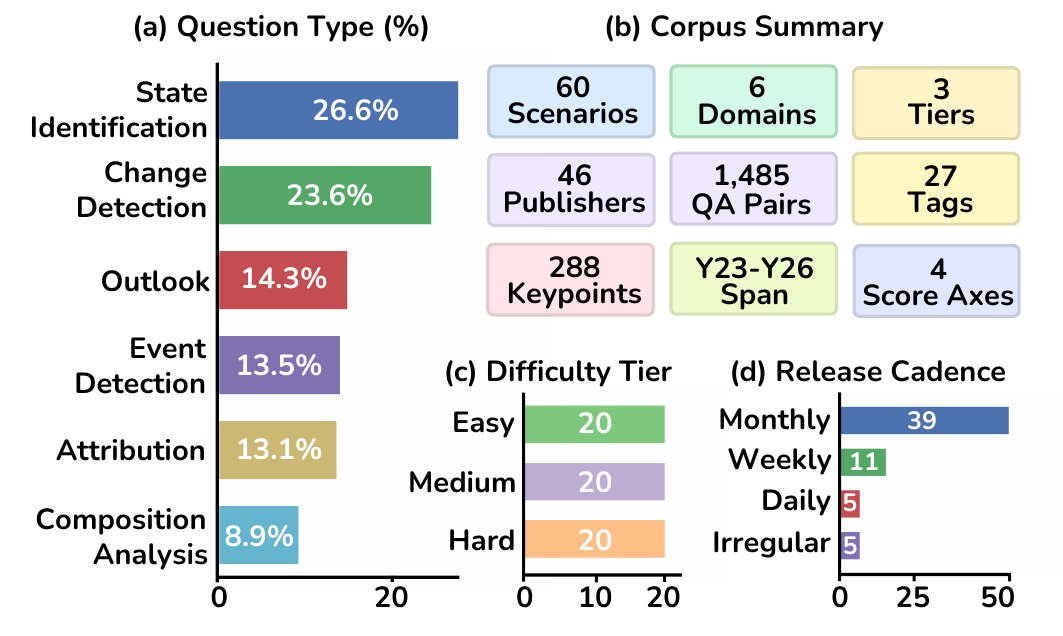}
\end{center}
\vspace{-11pt}
\caption{TimeSage-EV corpus statistics.}
\vspace{-10pt}
\label{Figure_3}
\end{figure}

\textbf{TimeSage-EV} is built from public institutional releases that recur on a fixed cadence, and every evaluation period of every scenario is reproducible from those releases. The pipeline proceeds in three stages: source-side data collection, scenario-side generation of period-level instances, and a quality control pass that enforces temporal isolation. Figure~\ref{Figure_3} reports corpus statistics, while Figure~\ref{Figure_4} sketches the pipeline. Full prompts, scenario checklists, and audit attestations are provided in Appendix~\ref{appendix_3}.

\subsubsection{Data Collection}

\begin{figure*}[htbp]
\begin{center}
\includegraphics[width = \linewidth]{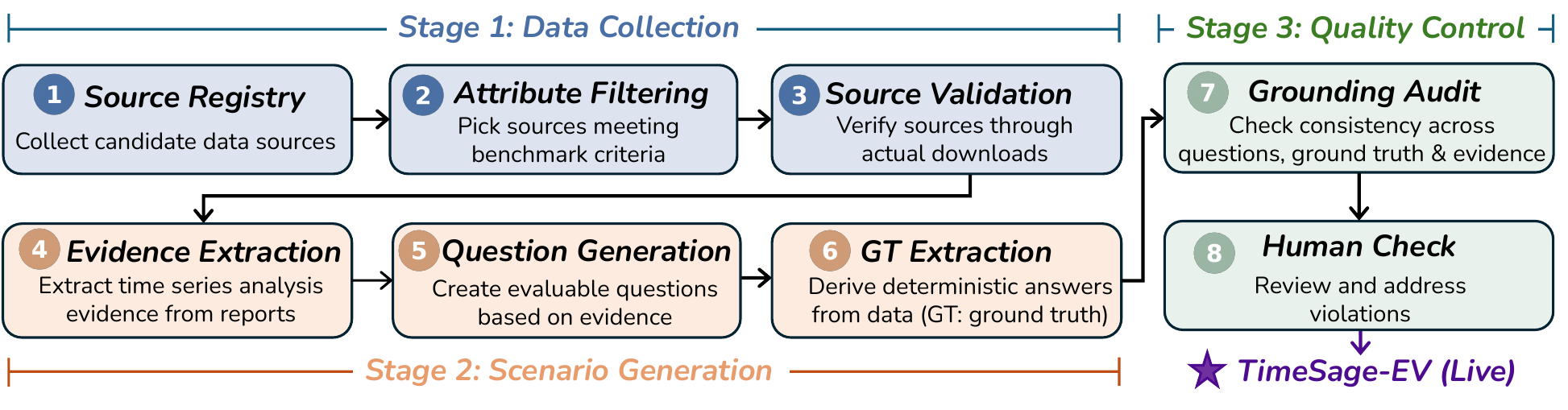}
\end{center}
\vspace{-8pt}
\caption{Overview of benchmark generation pipeline. Benchmark data is updated monthly via automated scripts.}
\vspace{-10pt}
\label{Figure_4}
\end{figure*}

\noindent\textbf{Step I: Domain scanning and source registry.}
We built a registry of 500 public institutional sources across six domains: agriculture, climate, energy, finance, healthcare, and transportation. Candidate sources include APIs, periodic briefs, and statistical reports from government agencies and international organizations. For each source, we record metadata, including publisher, source type, release cadence, historical span, and data format.

\noindent\textbf{Step II: Attribute-based filtering.}
We retain sources that are public, recurring, historically deep enough for repeated instantiation, grounded in timestamped time series, and tied to answers traceable to artifacts. We exclude sources requiring private access and lacking period alignment. We balance domain coverage, source type, and release cadence so that the corpus spans monthly, weekly, daily, and irregular schedules.

\noindent\textbf{Step III: Source validation.}
Each surviving source is validated for all target periods before promotion. We verify that source artifacts can be recovered, time series remain stable and ordered, cumulative histories can be assembled for each target period, and ground-truth fields are traceable to time series observations or quoted document spans. 

\subsubsection{Scenario Generation}

\noindent\textbf{Step IV: Evidence collection.}
For each accepted sources, we construct period-level evidence packages. Each package represents the agent-visible state at a given period, containing the time series and documents accumulated up to that point. We extract background information from the source description and time series-relevant analytical text from the documents. These materials provide the basis for subsequent question generation.

\noindent\textbf{Step V: Question generation.}
Using the evidence in Step IV, we generate questions and export a scenario-level contract for each source. To do this, we adapt question, answer-schema, and scoring templates from a 27-tag taxonomy, spanning six question categories (Appendix~\ref{app:sec_data_collection} Table~\ref{tab:question_taxonomy}), into a scenario-specific contract. LLMs are used for this adaptation, while the predefined question--answer--scoring structure is preserved. The contract can be instantiated in every period. The question distribution varies by difficulty: easy scenarios focus on state identification; medium scenarios require temporal pattern identification; and hard scenarios involve analytical reasoning that combines numerical changes with document-based explanations. This difficulty cliff is shown in Appendix~\ref{appendix:additional_results} Table~\ref{table_3}: keypoint answer accuracy declines monotonically from easy to hard across all evaluated LLMs.

\noindent\textbf{Step VI: Ground-truth extraction.}
For each instance, we derive reproducible ground truth according to the scenario contract. For each period, we follow the answer schema and scoring method specified in the scenario contract to construct ground truth, either through computation or LLM-assisted evidence finding, depending on the field type. For fields defined over the time series, we apply deterministic rules to the time series table, including threshold-based classification, multi-variable comparison, and cross-period comparison. For fields requiring textual evidence, we use the withheld target-period source during answer construction and verification, transcribing the statements or explanations from the document.

\subsubsection{Quality Control}
Every scenario undergoes an automated grounding audit and human review before release. The grounding audit checks schema validity, temporal consistency, variable and answer-space validity, deterministic answer derivability, evidence support, leakage prevention, and instance-level completeness. Three human reviewers then verify the ground truth, the clarity of the questions, and the grounding of the evidence. A scenario is released only after passing both stages. The final corpus contains 60 scenarios from 6 domains and 1{,}485 period-level instances spanning February 2023 to May 2026, with a balanced 20/20/20 difficulty-tier split, as summarized in Figure~\ref{Figure_3}.

\subsection{Evaluation Protocol}
Each agent run produces two outputs at each period: a structured answer and a free-form report. We score them on four axes: \textit{\textbf{(1)}} \textbf{keypoint accuracy}, which checks whether the structured answer matches the ground truth; \textit{\textbf{(2)}} \textbf{keypoint coverage}, which checks whether the report mentions the keypoints; \textit{\textbf{(3)}} \textbf{report faithfulness}, which checks whether the report's claims are supported by cutoff-valid evidence rather than future or unsupported information; and \textit{\textbf{(4)}} \textbf{report quality}, which checks whether the report is clear, organized, and appropriate for the scenario. The protocol combines rule-based scoring for fields with fixed answer formats, such as labels, yes/no values, and numbers, with LLM-based judging for the report's coverage, faithfulness, and writing quality. Each axis is reported on a 0--100 scale, and the run-level score is the average of the scored axes. We report both independent evaluation, where the agent starts fresh for each period, and sequential evaluation, where the agent can carry state within a scenario but receives only prior-period evidence. Figure~\ref{Figure_5} shows a scenario example with the evaluation protocol. More details on the judge are shown in Appendix~\ref{appendix_evaluation}.

\begin{figure}[htbp]
\begin{center}
\includegraphics[width = \linewidth]{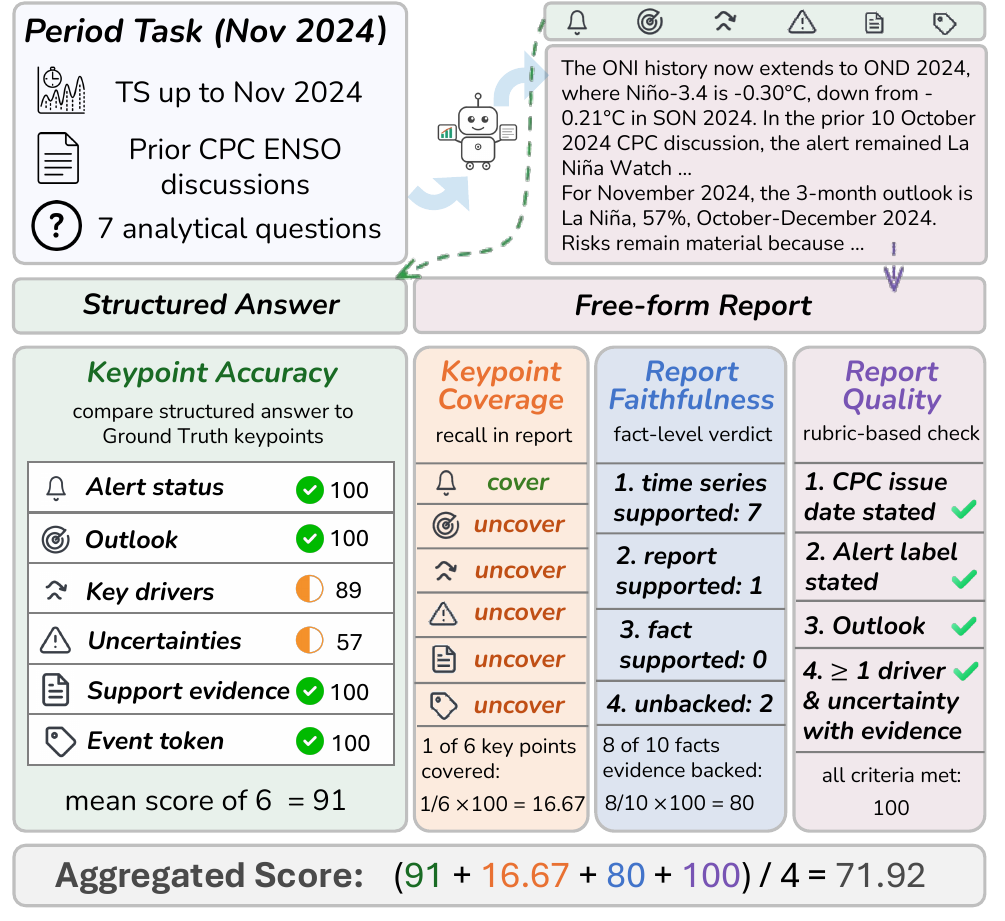}
\end{center}
\vspace{-8pt}
\caption{Evaluation for the El Niño–Southern Oscillation (ENSO) monitoring scenario. TS: time series.}
\vspace{-10pt}
\label{Figure_5}
\end{figure}

\section{Experiments}

\subsection{Environment Setting}

Each evolving scenario is instantiated as a sandbox that periodically releases time series analysis tasks and coordinates information exposure, evaluation, and feedback. In each period, the environment publishes the task and exposes only causally admissible information prior to that time point. The agent completes the task within this temporally isolated sandbox and submits its output. Once the scenario source releases the corresponding report, the environment collects the ground-truth outcome, evaluates the output, returns feedback, and advances to the next period, repeating the cycle. With a collection cutoff in May 2026, we snapshot the latest 24 periods per scenario, while an automated pipeline continuously pulls from the scenario sources so each environment updates at its own release cadence, and the benchmark remains live over time. All experiments in this paper use these 24 periods.

\subsection{Experimental Setting}

\noindent\textbf{Baseline Models.}
We use \texttt{smolagent}~\cite{smolagents} as the baseline, equipped with memory, search, and code execution. A memory-compression tool can be invoked by the agent or triggered automatically when the context approaches the token limit, as detailed in Appendix~\ref{appendix:experiment_setting}. We evaluate six LLMs under this harness: Sonnet-4.6~\cite{anthropic2026sonnet46}, GPT-5.4~\cite{openai2026gpt54}, Kimi-K2.6~\cite{moonshot2025kimik2}, Qwen-3.5-397B-A17B-NVFP4-TBC~\cite{qwen2026qwen35}, Devstral-2-123B-Instruct-2512~\cite{mistral2025devstral2}, and Gemma-4-31B-IT-NVFP4~\cite{google2026gemma4}.\footnote{Sonnet-4.6 and GPT-5.4 are accessed through the OpenRouter API (\url{https://openrouter.ai}), while the remaining LLMs are served via local vLLM deployments.} All models use  temperature=0 with a 32{,}768-token output budget; all other hyperparameters remain at provider defaults. Each rollout is limited to 20 steps or 2{,}700 seconds; runs that exhaust either budget without producing an answer are counted as failures. For the LLM judge axes (coverage, faithfulness, and quality), we use DeepSeek-V4-Flash~\cite{deepseekai2026deepseekv4}, with the prompts shown in Appendix~\ref{appendix:evaluation_prompts} and verification in Appendix~\ref{appendix:llmjudge_verification}.

\noindent\textbf{TimeSage-1.0.}
To examine whether agents can self-improve from feedback over long-horizon tasks, we propose TimeSage-1.0, a \texttt{smolagent}-based harness augmented with a self-evolving skill-library management system (see Appendix~\ref{appendix:experiment_setting} for more details). The system supports skill induction, quality checking, and skill retrieval, allowing agents to convert reasoning traces into reusable workflows and deploy them in subsequent periods.

\begin{figure*}[htbp]
\begin{center}
\includegraphics[width = \linewidth]{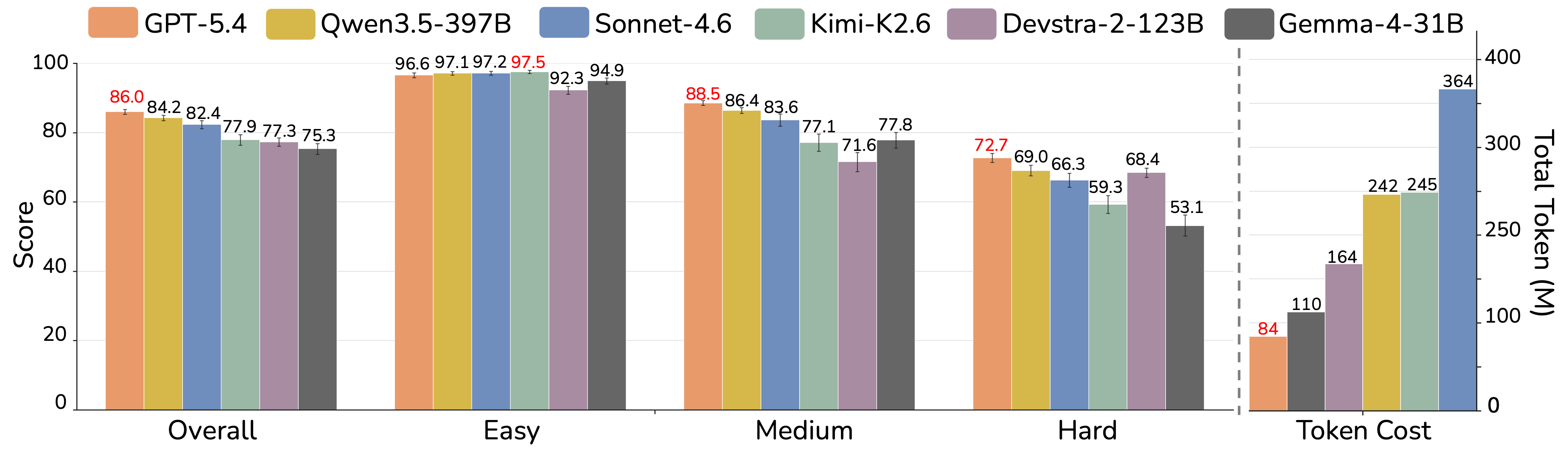}
\end{center}
\vspace{-10pt}
\caption{\textbf{Left}: Outcome scores across difficulty tiers and overall; error bars are per-difficulty-level percentile bootstrap 95\% confidence intervals. \textbf{Right}: Total token cost.}
\label{Figure_6}
\vspace{-6pt}
\end{figure*}

\definecolor{hc0}{RGB}{209,223,214}
\definecolor{hc1}{RGB}{208,222,213}
\definecolor{hc2}{RGB}{214,226,218}
\definecolor{hc3}{RGB}{228,236,231}
\definecolor{hc4}{RGB}{248,250,248}
\definecolor{hc5}{RGB}{249,251,249}
\definecolor{hc6}{RGB}{209,223,214}
\definecolor{hc7}{RGB}{205,220,211}
\definecolor{hc8}{RGB}{212,225,217}
\definecolor{hc9}{RGB}{222,232,225}
\definecolor{hc10}{RGB}{245,248,246}
\definecolor{hc11}{RGB}{248,250,248}
\definecolor{hc12}{RGB}{215,227,219}
\definecolor{hc13}{RGB}{214,226,219}
\definecolor{hc14}{RGB}{217,229,221}
\definecolor{hc15}{RGB}{233,240,235}
\definecolor{hc16}{RGB}{250,252,251}
\definecolor{hc17}{RGB}{255,255,255}
\definecolor{hc18}{RGB}{208,222,213}
\definecolor{hc19}{RGB}{204,220,210}
\definecolor{hc20}{RGB}{213,225,217}
\definecolor{hc21}{RGB}{223,232,226}
\definecolor{hc22}{RGB}{245,248,246}
\definecolor{hc23}{RGB}{247,250,248}
\definecolor{hc24}{RGB}{215,227,220}
\definecolor{hc25}{RGB}{221,231,224}
\definecolor{hc26}{RGB}{218,229,222}
\definecolor{hc27}{RGB}{230,238,233}
\definecolor{hc28}{RGB}{244,248,246}
\definecolor{hc29}{RGB}{248,250,248}
\definecolor{hc30}{RGB}{216,228,220}
\definecolor{hc31}{RGB}{218,229,222}
\definecolor{hc32}{RGB}{220,230,224}
\definecolor{hc33}{RGB}{237,243,239}
\definecolor{hc34}{RGB}{252,253,252}
\definecolor{hc35}{RGB}{254,254,254}

\definecolor{sg1}{RGB}{244,247,245}
\definecolor{sg2}{RGB}{226,235,229}
\definecolor{sg3}{RGB}{205,221,212}
\definecolor{sg4}{RGB}{180,202,189}
\newcommand{\scalebar}{%
  \raisebox{-0.1ex}{\setlength{\fboxsep}{0pt}%
  \colorbox{sg1}{\rule{0pt}{1.6ex}\hspace{1.6ex}}%
  \colorbox{sg2}{\rule{0pt}{1.6ex}\hspace{1.6ex}}%
  \colorbox{sg3}{\rule{0pt}{1.6ex}\hspace{1.6ex}}%
  \colorbox{sg4}{\rule{0pt}{1.6ex}\hspace{1.6ex}}}}

\begin{table*}[t]
\centering
\caption{Per-question-type scores, reported as mean with bootstrapped 95\% CI in the format $x^{+\mathrm{high}}_{-\mathrm{low}}$. Cell shading uses a shared scale
(low~\scalebar~high); the best result in each column is shown in bold.}
\label{table_2}
\vspace{-5pt}
\setlength{\tabcolsep}{6pt}\small
\renewcommand{\arraystretch}{1.5}
\begin{tabularx}{\textwidth}{@{}l>{\centering\arraybackslash}X>{\centering\arraybackslash}X>{\centering\arraybackslash}X>{\centering\arraybackslash}X>{\centering\arraybackslash}X>{\centering\arraybackslash}X@{}}
\toprule
\textbf{Model} & \textbf{State ID} & \textbf{Change Det.} & \textbf{Comp. ID} & \textbf{Event Det.} & \textbf{Attribution} & \textbf{Forecast} \\
\midrule
\logo{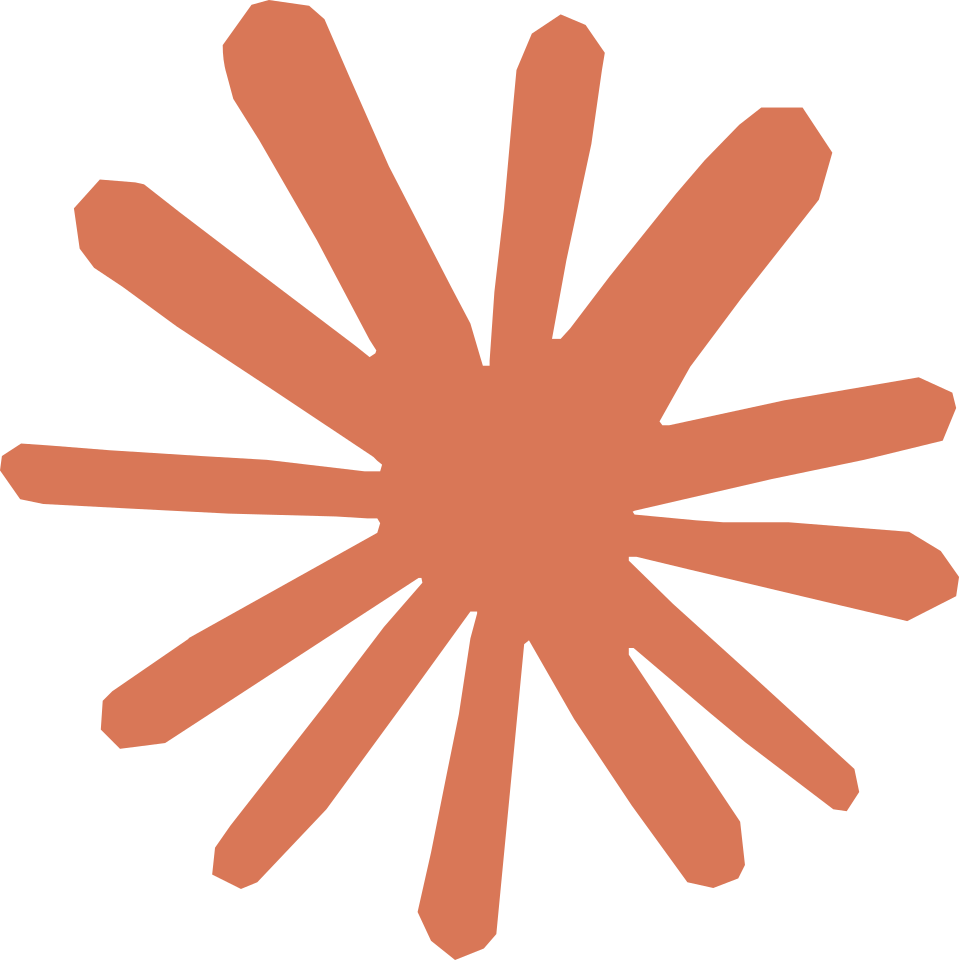}Sonnet-4.6 & \cellcolor{hc0}\val{90.9}{+1.2}{-1.2} & \cellcolor{hc1}\val{92.9}{+1.5}{-1.7} & \cellcolor{hc2}\val{85.6}{+2.7}{-2.7} & \cellcolor{hc3}\val{67.9}{+2.2}{-2.1} & \cellcolor{hc4}\val{42.6}{+3.0}{-3.2} & \cellcolor{hc5}\val{41.2}{+4.1}{-4.0} \\
\logo{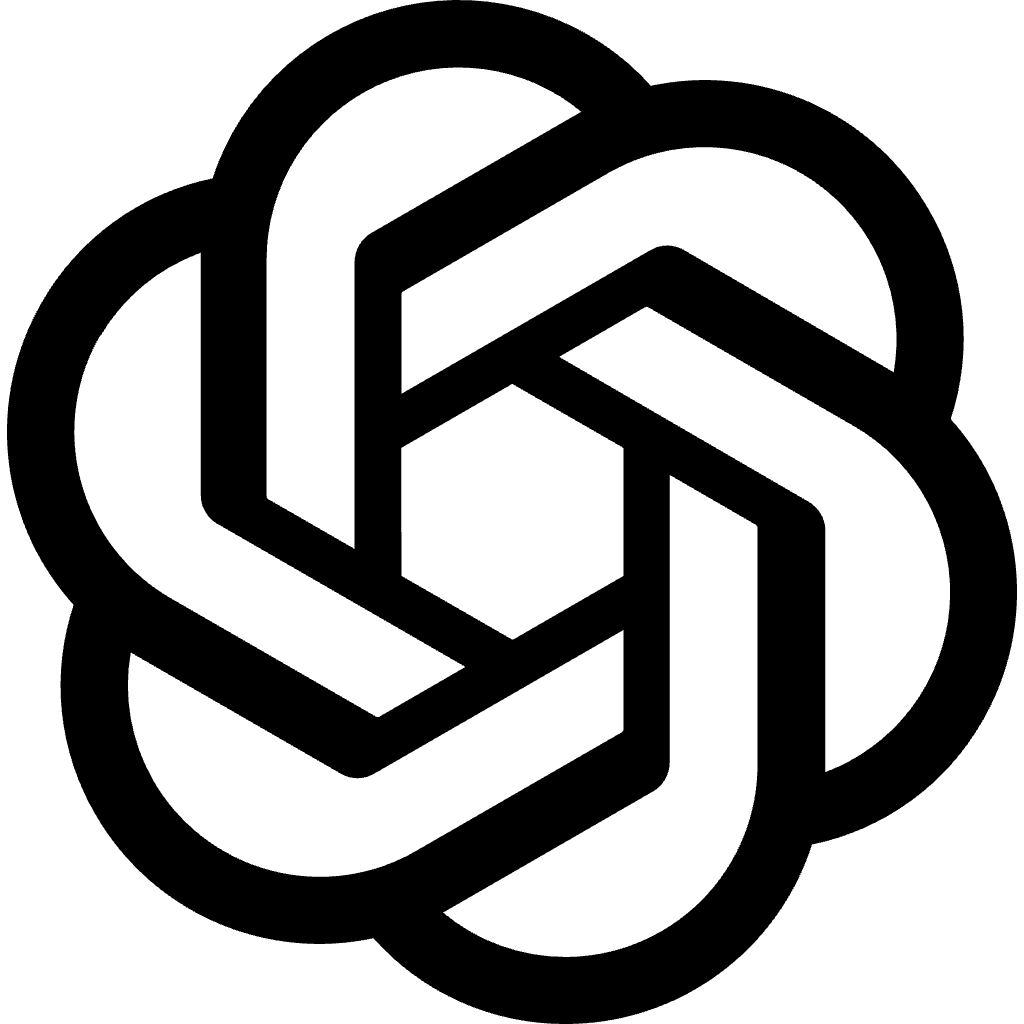}GPT-5.4 & \cellcolor{hc6}\valb{93.8}{+0.8}{-0.9} & \cellcolor{hc7}\val{96.0}{+1.2}{-1.2} & \cellcolor{hc8}\valb{86.9}{+2.4}{-2.7} & \cellcolor{hc9}\valb{75.4}{+1.9}{-1.9} & \cellcolor{hc10}\val{46.2}{+3.2}{-3.4} & \cellcolor{hc11}\val{42.7}{+4.1}{-3.9} \\
\logo{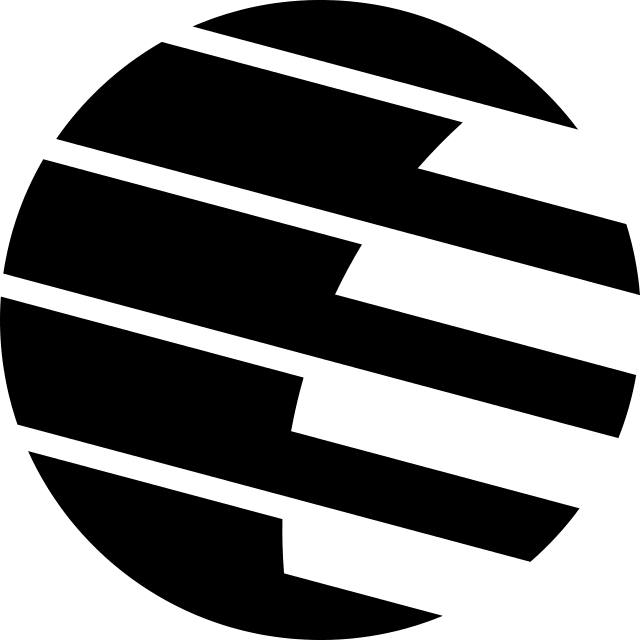}Kimi-K2.6 & \cellcolor{hc12}\val{84.2}{+1.8}{-1.7} & \cellcolor{hc13}\val{84.9}{+2.4}{-2.4} & \cellcolor{hc14}\val{80.8}{+3.2}{-3.1} & \cellcolor{hc15}\val{61.0}{+2.6}{-2.5} & \cellcolor{hc16}\val{39.3}{+3.3}{-3.4} & \cellcolor{hc17}\val{33.4}{+3.9}{-3.7} \\
\logo{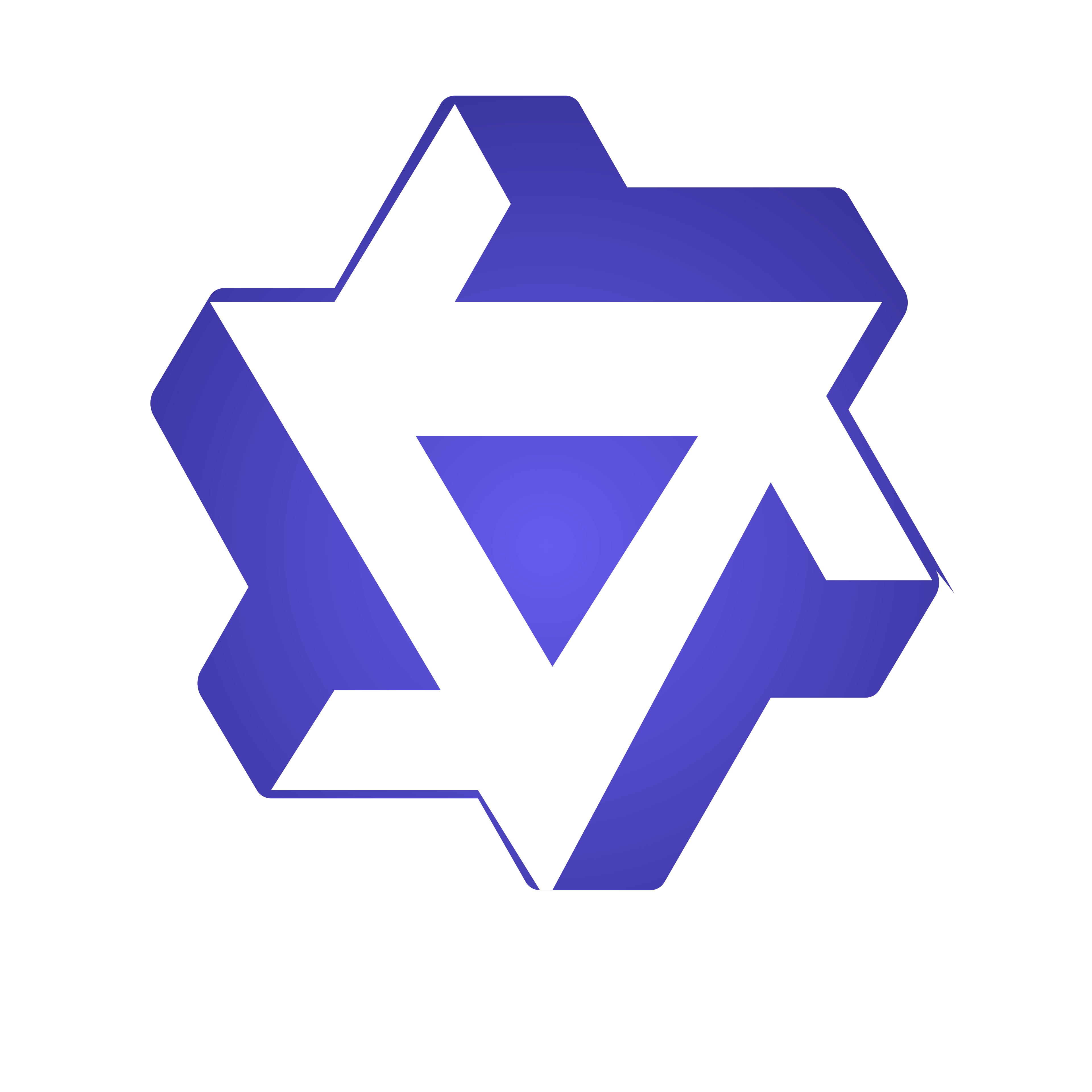}Qwen-3.5-397B & \cellcolor{hc18}\val{92.6}{+0.9}{-0.9} & \cellcolor{hc19}\valb{97.0}{+0.9}{-1.0} & \cellcolor{hc20}\val{86.8}{+2.4}{-2.5} & \cellcolor{hc21}\val{74.1}{+1.9}{-1.9} & \cellcolor{hc22}\val{46.1}{+3.3}{-3.2} & \cellcolor{hc23}\valb{43.0}{+4.0}{-3.8} \\
\logo{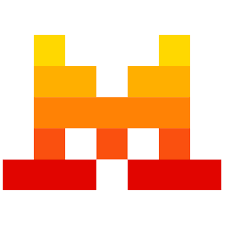}Devstral-2-123B & \cellcolor{hc24}\val{83.4}{+1.5}{-1.6} & \cellcolor{hc25}\val{76.5}{+2.8}{-2.6} & \cellcolor{hc26}\val{79.7}{+3.1}{-3.2} & \cellcolor{hc27}\val{64.7}{+2.4}{-2.4} & \cellcolor{hc28}\valb{46.7}{+3.2}{-3.0} & \cellcolor{hc29}\val{42.8}{+4.1}{-4.3} \\
\logo{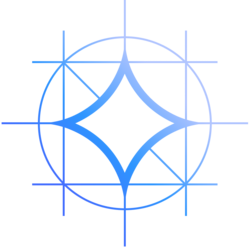}Gemma-4-31B & \cellcolor{hc30}\val{82.7}{+1.7}{-1.8} & \cellcolor{hc31}\val{80.2}{+2.4}{-2.7} & \cellcolor{hc32}\val{77.3}{+3.4}{-3.3} & \cellcolor{hc33}\val{55.7}{+2.7}{-2.6} & \cellcolor{hc34}\val{37.3}{+3.4}{-3.2} & \cellcolor{hc35}\val{34.8}{+4.0}{-3.8} \\
\bottomrule
\end{tabularx}
\vspace{-10pt}
\end{table*}

\subsection{Main Results}
\noindent\textbf{Overall performance.}
Across the full benchmark, GPT-5.4 achieves the strongest score (86.0), followed by Qwen-3.5-397B (84.2) and Sonnet-4.6 (82.4), as summarized in Figure~\ref{Figure_6}. Performance is near-saturated on easy tasks, where all models score above 92, but drops as the task requires richer analytical synthesis. On hard tasks, the best model reaches only 72.7, while Kimi-K2.6 and Gemma-4-31B fall to 59.3 and 53.1, respectively.
This tier-wise gap indicates that agents can handle basic time series analysis tasks, including state and temporal pattern identification. However, \textit{large-scale evidence extraction and external-information reasoning remain major bottlenecks in hard scenarios.}

\noindent\textbf{Token cost.}
As shown on the right side of Figure~\ref{Figure_6}, total token cost varies by more than 4$\times$ across models, from 84M for GPT-5.4 to 364M for Sonnet-4.6. GPT-5.4 achieves the highest overall score while using the fewest tokens, showing a clear efficiency advantage. In contrast, Sonnet-4.6 and other high-cost models fail to lead in accuracy. These results suggest that \textit{reasoning efficiency drives performance in long-horizon time series analysis.}

\begin{figure}[htbp]
\begin{center}
\includegraphics[width = \linewidth]{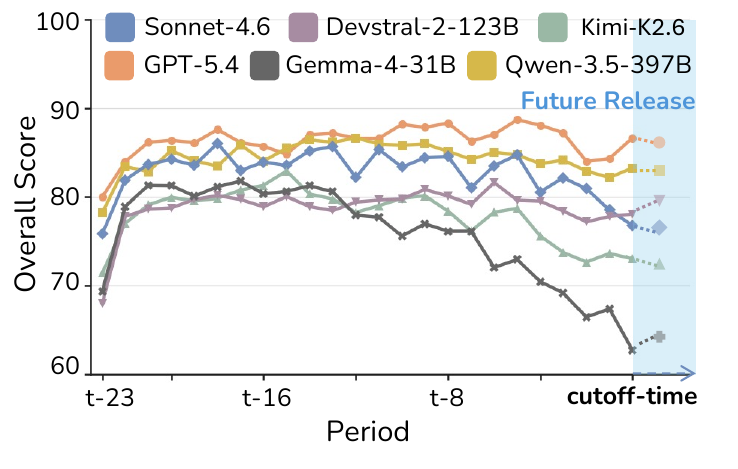}
\end{center}
\vspace{-12pt}
\caption{Live overall-score leaderboard across periods. Scores are tracked as information accumulates from $t-23$ to cutoff time. Shaded region marks future updates.}
\label{Figure_7}
\vspace{-10pt}
\end{figure}

\noindent\textbf{Live leaderboard.}
Figure~\ref{Figure_7} traces how overall scores evolve under continuous testing, where each LLM's in-context history accumulates from period $t-23$ to the cutoff time. After an initial three-period ramp-up, all models reach a high plateau, indicating that early context is useful. The trajectories then diverge: GPT-5.4 maintains the highest and most stable score with no degradation, showing strong long-horizon memory robustness, whereas Gemma-4-31B and Kimi-K2.6 decline steadily, the former falling from 80 to nearly 63. This late-stage drop suggests that \textit{long-horizon history retrieval becomes a bottleneck for weaker LLMs in time series analysis}. This is a key advantage of \ours{} over static benchmarks: rather than evaluating isolated snapshots, it exposes whether LLMs remain reliable under long-term deployment.

\noindent\textbf{Per-question-type analysis.}
The per-question-type results in Table~\ref{table_2} reveal where the difficulty arises across the six question families. State identification, change detection, and composition questions are comparatively tractable, with the strongest models reaching above 85--95. In contrast, attribution and forecast questions remain difficult for all models. The best attribution score is 46.7 and the best forecast score is 43.0. These results reveal a capability gap: \textit{agents can compute over time series, but struggle with temporal grounding for cautious forward-looking analysis.}

\subsection{Failure Mode (FM) Analysis}

\begin{figure}[htbp]
\vspace{-5pt}
\begin{center}
\includegraphics[width = \linewidth]{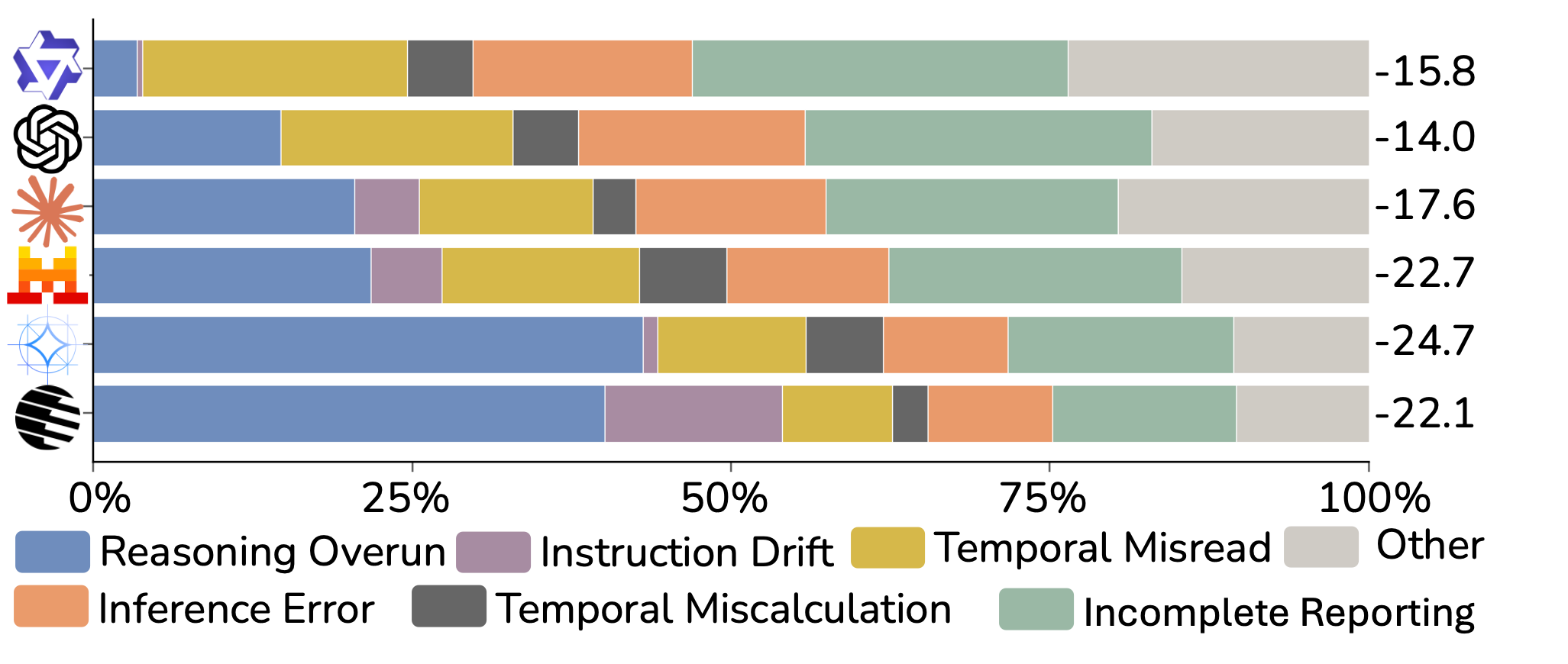}
\end{center}
\vspace{-10pt}
\caption{Failure-mode distribution per model; right numbers indicate total score impact.}
\label{Figure_8}
\vspace{-10pt}
\end{figure}

\noindent\textbf{FM1: Reasoning failure.}
Figure~\ref{Figure_8} shows that reasoning failures are a major failure mode across models, involving reasoning overrun and instruction drift. Reasoning overrun occurs when a model follows a long reasoning path and times out, reflecting difficulty reaching the correct answer efficiently. Instruction drift occurs when long contexts cause the model to forget rules, such as code-execution protocols or bash-command constraints. These failures highlight the need for efficient reasoning and memory management in evolving environments.

\noindent\textbf{FM2: Temporal misuse.}
Temporal errors, including temporal misreading, temporal miscalculation, and inference errors stems from a weak alignment between task semantics, time series data, and evidence. Agents may read incorrect rows or timestamps, compute incorrect variables or time windows, or rely on mismatched evidence for forecasting and attribution. These errors show that temporal grounding, rather than data access, remains a core challenge in evolving time series analysis environments.

\noindent\textbf{FM3: Incomplete reporting.}
Incomplete reporting is a failure mode, where agents recover correct structured fields but omit required key points or fail to ground them in the report. This occurs in part because overly long reasoning chains make it difficult for agents to summarize all required points.

\subsection{The Roles of Memory and Skill Induction}

\begin{figure}[htbp]
\begin{center}
\includegraphics[width = \linewidth]{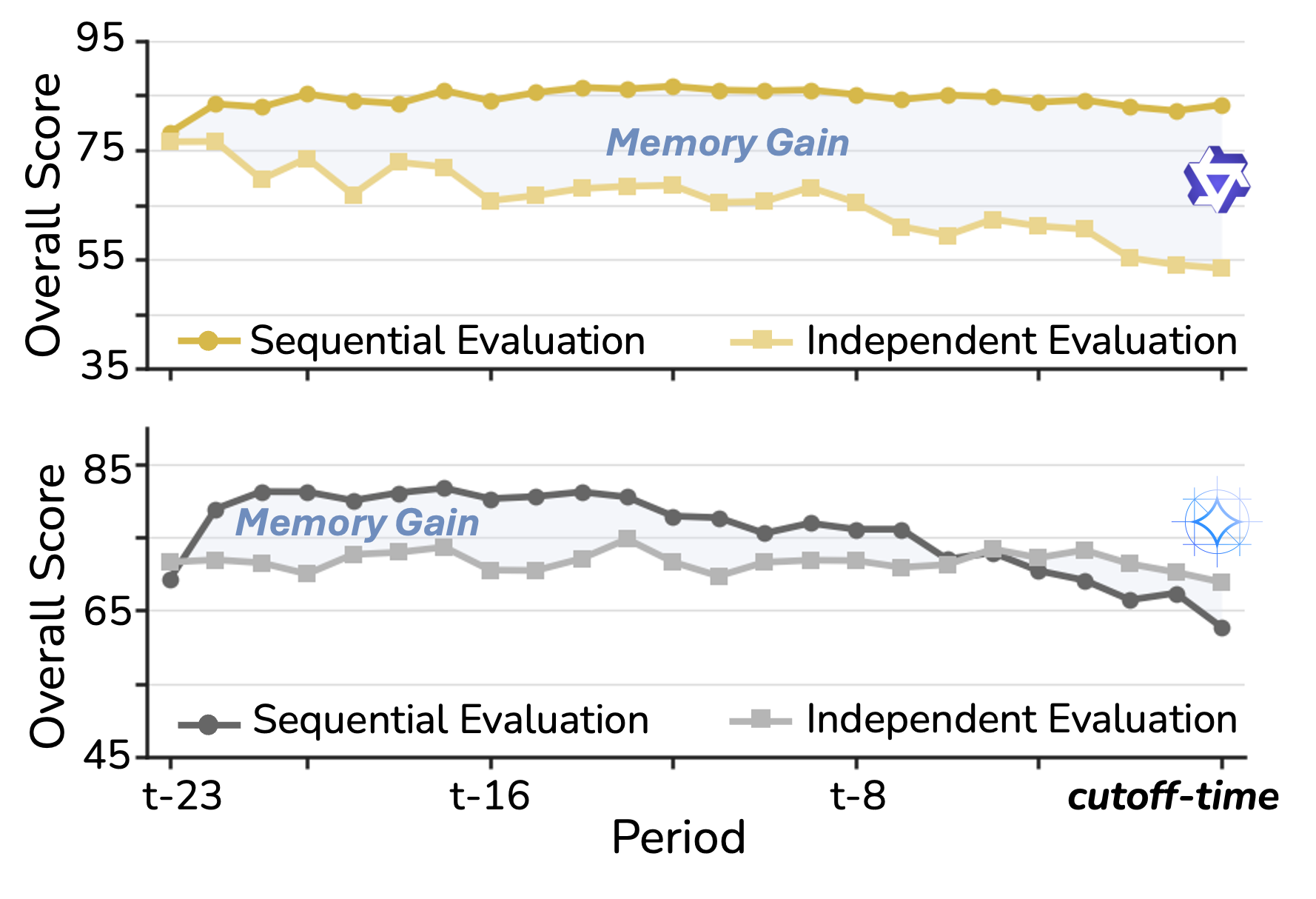}
\end{center}
\vspace{-10pt}
\caption{Sequential vs. independent evaluation for Qwen-3.5-397B (Top) and Gemma-4-31B (Bottom).}
\label{Figure_9}
\vspace{-5pt}
\end{figure}

\paragraph{Does Memory Help in Evolving Environments?}
Figure~\ref{Figure_9} compares the sequential and independent evaluation of Qwen-3.5-397B and Gemma-4-31B. For Qwen-3.5-397B, memory provides a benefit: sequential evaluation outperforms independent evaluation over periods. This suggests that a persistent state helps agents reuse analysis patterns instead of solving each period from scratch. However, for Gemma-4-31B, memory is less beneficial: sequential evaluation improves early performance, but its advantage narrows and reverses in later periods. \textit{Memory is useful only when models can manage it: history can support future analysis but may become a burden for in-context management.}

\begin{figure}[htbp]
\begin{center}
\includegraphics[width = \linewidth]{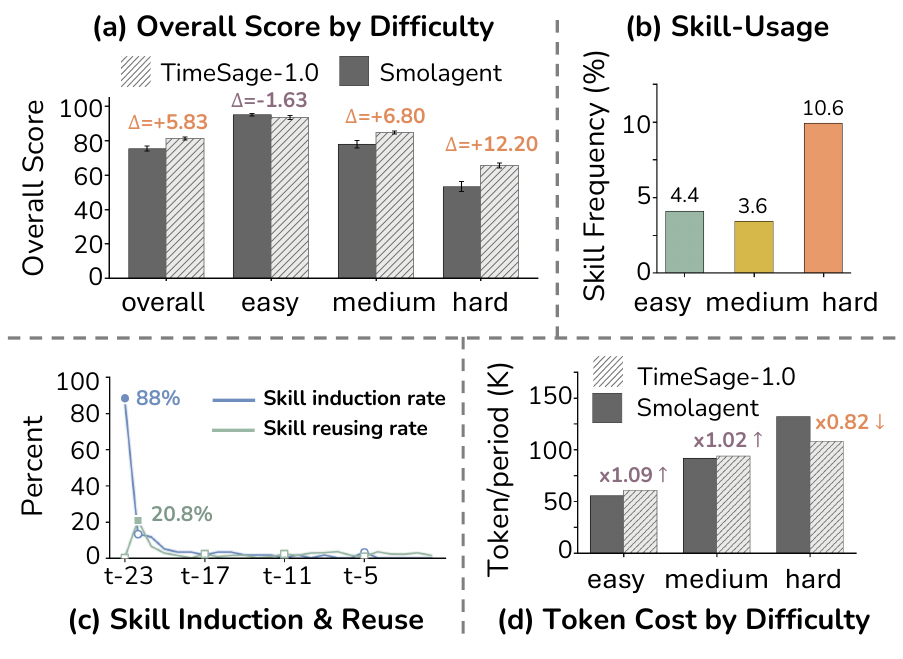}
\end{center}
\vspace{-8pt}
\caption{Analysis of self-evolving skill mechanism on Gemma-4-31B under evolving evaluation environments.}
\label{Figure_10}
\vspace{-5pt}
\end{figure}

\paragraph{Can Agents Self-Evolve Reusable Skills?}
To test whether agents can convert repeated feedback into reusable analytical skills, we evaluate \textbf{TimeSage-1.0}, our \texttt{smolagent}-based harness equipped with a lightweight self-evolving skill-library manager. As shown in Figure~\ref{Figure_10} a, reusable skills improve overall performance, with the clearest gains in medium- and hard-difficulty scenarios. To understand this benefit, we further examine efficiency and skill usage: enabling reusable skills reduces token cost to $0.82\times$ (Figure~\ref{Figure_10} d), while the skill-invocation frequency peaks in hard scenarios (Figure~\ref{Figure_10} b), suggesting that \textit{reusable skills are most valuable in evolving time series analysis when reasoning is difficult}. However, skill induction and reuse are concentrated in the early periods and do not persist across the entire horizon (Figure~\ref{Figure_10} c), revealing that maintaining an effective skill library remains challenging in evolving time series tasks.

\section{Conclusion}
We introduced \textbf{TimeSage-EV}, a live benchmark for \textbf{agentic time series analysis in evolving environments}. TimeSage-EV formalizes recurring, cutoff-valid analytical updates and evaluates agents across 60 institutional scenarios, 1,485 scenario-period QA pairs, and 4 complementary scoring axes. Experiments with frontier LLM agents and TimeSage-1.0 reveal persistent gaps in memory management, long-horizon reasoning, and temporal validity. These findings establish TimeSage-EV as a diagnostic benchmark for developing robust analytical agents that can update time series analyses reliably as evidence continually evolves.

\section*{Limitations}

TimeSage-EV is built from public institutional sources with recurring releases, and therefore does not cover enterprise data, dashboards, or settings where historical releases are unavailable or poorly archived. Although the benchmark spans 6 domains and multiple release cadences, it cannot exhaust the diversity of document formats, institutional conventions, or analytical workflows. Furthermore, the dataset relies predominantly on English-language reports from established global institutions, which may introduce geographic and linguistic biases, and it focuses strictly on numerical time series and textual documents, excluding the visual interpretation of charts and graphs that frequently accompany real-world releases. 

Regarding our evaluation protocol, we combine rule-based scoring for structured fields with LLM-based judging for report coverage, faithfulness, and quality. Despite structured rubrics and evidence grounding, some residual judge noise and bias may remain. Our experimental baseline also relies on the \texttt{smolagent} framework, meaning some observed failure modes, such as instruction drift or reasoning overrun, may be partially entangled with this specific harness rather than solely reflecting the underlying capabilities of the LLMs. 

The experiments in this paper use May 2026 snapshot. We will release a frozen, versioned snapshot of the benchmark containing all 1,485 scenario-period instances evaluated in this paper (capped at May 2026). Future researchers can evaluate against this static version for direct comparability with our baselines, while simultaneously testing on the live branch to measure true out-of-distribution generalization on newly published institutional releases.

\section*{Intended Use}
TimeSage-EV is intended as an academic research benchmark for diagnosing LLM agents on recurring time series analysis, and agent output should be treated as decision-support signals rather than substitutes for professional judgment in high-stakes domains. Scores in TimeSage-EV should be interpreted as diagnostic indicators of analytical behavior in covered scenarios rather than as guaranties of agent reliability in unseen sources, domains, or release schedules. Agents may also produce fluent but unsupported claims or violate temporal cutoffs in ways that appear authoritative, so outputs require human verification before any downstream use.

\section*{Ethical Considerations}
TimeSage-EV strictly adheres to ethical research standards. It utilizes publicly available institutional releases in compliance with each source's usage terms. The benchmark contains only aggregate time series and reports, with no personally identifiable information (PII). Human reviewers verifying ground-truth correctness worked only with public, non-sensitive content. We emphasize that our agents' analyses are designed as a decision-support tool for human experts, not a replacement for professional judgment in high-stakes domains, and users should be mindful of potential model hallucinations. Our approach does not involve PII or sensitive data, and we follow the Code of Ethics.

\section*{Acknowledgments}

This work was financially supported by the European Union's Horizon Europe research and innovation programme under grant agreement No.~101214398 (ELLIOT). Views and opinions expressed are however those of the author(s) only and do not necessarily reflect those of the European Union or the European Commission. Neither the European Union nor the European Commission can be held responsible for them. 
We also acknowledge the Supercomputing Center of the Eindhoven University of Technology (\href{https://supercomputing.tue.nl/}{Home - TU/e Supercomputing Center}) for providing access to and assistance with the various computing resources available.

\begin{figure}[h]
    \vspace{-10pt}
    \centering
    \includegraphics[width=0.6\linewidth]{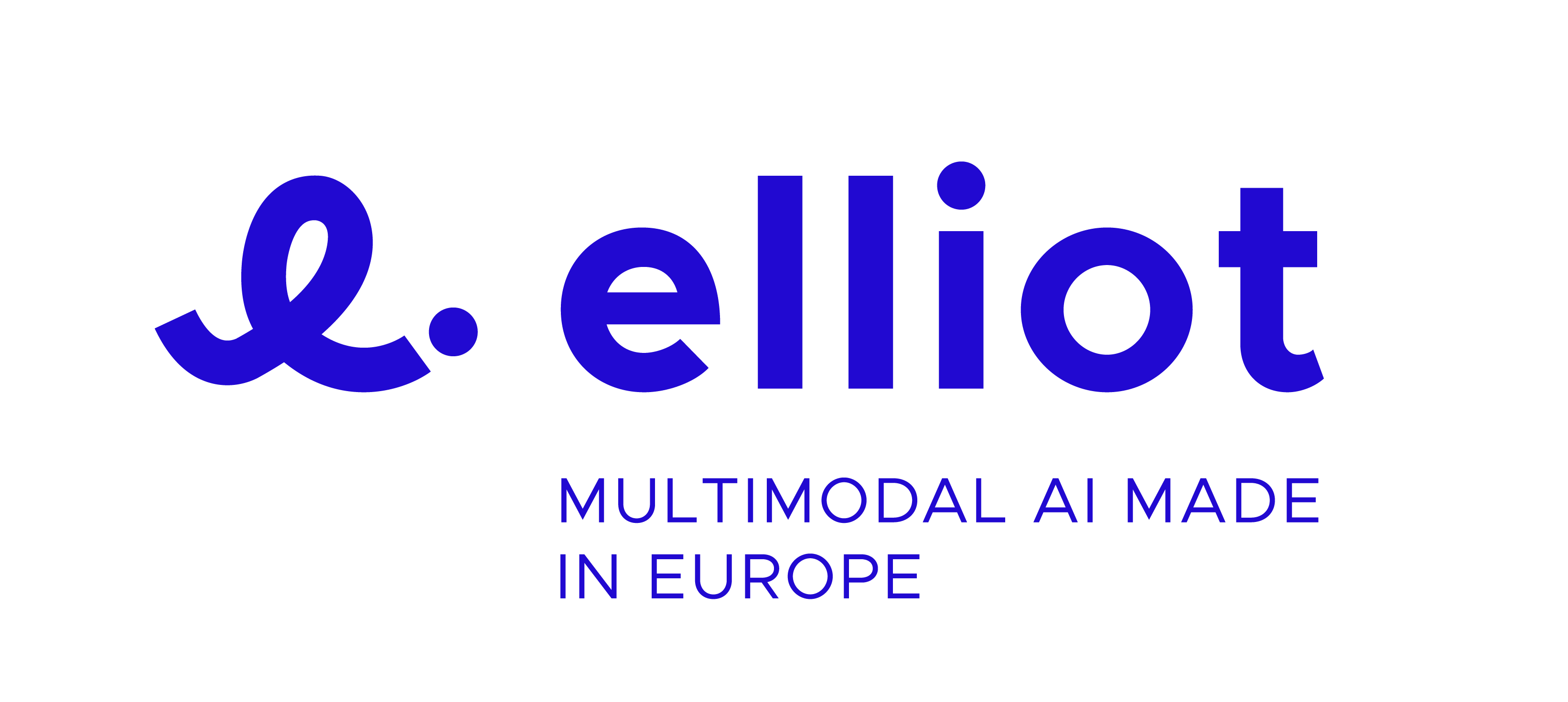}
\end{figure}

\newpage
\bibliography{reference}

\clearpage
\appendix

\section{Appendix}
\subsection{Task Rationale}
\label{appendix_1}

\paragraph{Why recurring time series analysis is different.}
TimeSage-EV targets a setting that is not captured by standard time series forecasting, static time series QA, or generic agent benchmarks. In recurring institutional analysis, the task specification remains fixed, but the evidence available to the analyst changes after each release. The agent must interpret the latest observation, compare it with historical context, use only cutoff-valid documents, and revise its report as the environment evolves.

\paragraph{Why temporal isolation matters.}
A central motivation of TimeSage-EV is to evaluate whether agents can reason within the information limit that would have been applied at a target period. Without this constraint, an agent may answer using future reports, post-hoc explanations, or documents published after the target release. TimeSage-EV prevents this by exposing cumulative time series data and prior-period source documents while withholding the target-period release for ground truth construction and evidence verification.

\paragraph{Why agentic evaluation matters.}
Real analytical workflows require more than reading a fixed table. Agents must inspect data files, use tools, consult documents, preserve state across periods, and produce both structured answers and natural-language reports. TimeSage-EV evaluates agentic time series analysis as an integrated workflow, exposing failures such as unsupported causal claims, poor use of evidence, and incomplete reporting.

\subsection{Core Concepts}
\label{appendix_2}

\paragraph{Live benchmark.}
We call TimeSage-EV a live benchmark because its evaluation set continuously expands as institutional releases are published. For each scenario $s$, new releases extend $\mathcal{T}_s$ over time. Each extension updates the cumulative time series and the available prior-period evidence, withholds the target-period release, and keeps the task contract fixed. This expansion is automated rather than manual: each scenario carries a self-contained, re-runnable pipeline that re-pulls the source, rebuilds the cumulative inputs, withholds the newest release, and re-derives ground truth and evidence under the scenario's frozen rules. A monthly maintenance script executes this pipeline across all scenarios, keeping TimeSage-EV live without re-authoring any questions and ground-truths.

\paragraph{Evolving environment.}
An evolving environment is a sequence of period-specific evidence for the recurring analytical task. The domain background, question template, and answer schema remain fixed within a scenario, while the time series extends, and the document set changes as the releases arrive. The agent must update its judgments for periods while respecting the cutoff point.

\paragraph{Agentic time series analysis.}
Agentic time series analysis refers to workflows in which an agent uses tools to inspect time series data, consult external evidence, answer analytical questions, and produce task-specific output. TimeSage-EV studies a distinctive setting in which the analytical task recurs across an evolving environment: the time series extends, the document set changes, and each period imposes a cutoff on evidence. The agent can maintain evidence, reasoning traces, or reusable analysis artifacts over periods. We therefore evaluate not only the structured answer, but also a free-form report that exposes the agent's evidence use, temporal grounding, and reasoning. This mirrors real-world time series analysis, where conclusions are typically delivered as periodic analytical reports rather than standalone structured predictions.

\subsection{Benchmark Construction}
\label{appendix_3}

Section~\ref{sec:benchmark_construction} and Figure~\ref{Figure_4} summarize the benchmark generation pipeline. This appendix expands that overview into a detailed construction protocol. Sections~\ref{app:sec_data_collection} through~\ref{app:sec_human_annotations} walk through the pipeline
sequentially, from data collection and scenario generation to quality control and human annotation. Section~\ref{appendix_data_sources} documents the data sources and reports volatility statistics. Finally,
Section~\ref{app:sec_scenario_examples} presents three illustrative scenario examples.
 


\subsubsection{Stage 1: Data Collection}
\label{app:sec_data_collection}

\paragraph{Step 1: Source registry.}
The pipeline is a domain-by-domain pass loop. Each domain pass begins from a target domain selected to close a current portfolio gap, such as an under-represented tier or release cadence. For that domain, the builder constructs a registry of candidate public sources, searching for recurring institutional data products that could fill the gap. We prioritize government agencies, central banks, international organizations and stable institutional publishers, including sources such as EIA, NOAA, CDC, FRED, BLS, the World Bank and IEA. Each candidate is recorded in eight standardized metadata fields: endpoint, authentication requirement, update frequency, historical span, data format, license, report links, and free-form notes. The step yields a per-domain candidate registry and exits when the domain has accumulated at least ten plausible candidates or is exhausted, with every metadata field either filled in or explicitly marked unknown.

\paragraph{Step 2: Attribute filtering.}
Taking the candidate registry as input, we first reject sources that cannot support a recurring benchmark task before writing any concrete question, and then assign a difficulty tier to the passed sources.
\begin{itemize}[leftmargin=*,itemsep=0pt,topsep=2pt]
    \item Metadata filtering: apply Table~\ref{tab:validity_filter} to remove sources that are inaccessible, weakly time-indexed, insufficiently variable, or unsuitable for evaluation of evolving-agents.
    \item Tier criteria: assign each surviving source to preliminary tiers based on Table~\ref{tab:tier_criteria}: \textit{Easy} for categorical state identification, \textit{Medium} for temporal-pattern summarization, and \textit{Hard} for report-grounded reasoning about causes or outlooks. 
\end{itemize}

\begin{table}[htbp]
\begin{center}
\captionof{table}{Metadata-level validity checks in Step 2.}
\label{tab:validity_filter}
\small
\renewcommand{\arraystretch}{1.18}
\resizebox{\columnwidth}{!}{
\begin{tabular}{p{0.045\columnwidth} p{0.895\columnwidth}}
\toprule
\textbf{No.} & \textbf{Validity check} \\
\midrule
1 & \textbf{Public access.} Source is publicly accessible without credentials, payment, or approval. \\
2 & \textbf{Time series grounding.} Task uses an observable timestamped series with values for the queried period. \\
3 & \textbf{Query persistence.} The same template applies to both historical and future periods. \\
4 & \textbf{Temporal anchoring.} The source provides a date, period, timestamp, or update window. \\
5 & \textbf{GT freshness.} Ground truth updates at least monthly, except report-level hard tasks. \\
6 & \textbf{Temporal variability.} Outputs vary over time and avoid near-constant majority labels. \\
7 & \textbf{Evolving-agent fit.} Task requires time-dependent retrieval, update, comparison, or synthesis. \\
\bottomrule
\end{tabular}}
\renewcommand{\arraystretch}{1.0}
\end{center}
\end{table}

\begin{table}[htbp]
\begin{center}
\captionof{table}{Difficulty-tier criteria across source, task, answer format, and evaluation in Step 2.}
\label{tab:tier_criteria}
\small
\renewcommand{\arraystretch}{1.18}
\resizebox{\columnwidth}{!}{
\begin{tabular}{p{0.1\columnwidth} p{0.78\columnwidth}}
\toprule
\textbf{Tier} & \textbf{Criteria} \\
\midrule
\textbf{Easy}
& \textbf{Source:} a data dashboard reporting current operational state as categorical indicators.
\textbf{Task:} identify the current state (reasoning LOW, capability dependence LOW).
\textbf{Format:} a single categorical label.
\textbf{Eval:} exact match against the correct option. \\
\addlinespace[3pt]
\textbf{Medium}
& \textbf{Source:} a data brief with time series and descriptive commentary, but no causal analysis.
\textbf{Task:} summarize recent changes across time points and sub-components (reasoning MEDIUM, dependence MEDIUM).
\textbf{Format:} a structured schema (period value, direction, leading sub-component, short summary).
\textbf{Eval:} LLM judge or keyword matching for per-field coverage and factual consistency. \\
\addlinespace[3pt]
\textbf{Hard}
& \textbf{Source:} an analytical report with trend judgements, drivers, risks, and outlooks.
\textbf{Task:} reason about causes of change and what comes next (reasoning HIGH, dependence HIGH).
\textbf{Format:} a structured schema (conclusion, key drivers, risks, uncertainties, evidence).
\textbf{Eval:} LLM judge or keyword matching, with each claim traceable to a section, table, or figure. \\
\bottomrule
\end{tabular}}
\renewcommand{\arraystretch}{1.0}
\end{center}
\end{table}

\paragraph{Step 3: Execution validation.}
The third step takes each retained source--tier pair and verifies by execution so that it can produce valid period-level benchmark instances. The builder retrieves the primary time series, releases archive, normalizes it into a period-indexed table, and retrieves the corresponding source releases, such as dashboards, briefs, bulletins, or analytical reports. As summarized in Table~\ref{tab:execution_validation}, each candidate must provide at least \(24\) periods aligned instances with traceable, time-isolated, and rule-evaluable evidence.

The difficulty tier assigned in Step 2 is confirmed during execution: \textit{Easy} scenarios must yield categorical state-identification instances, \textit{Medium} scenarios must yield summaries of temporal patterns, and \textit{Hard} scenarios must yield report-grounded reasoning about causes or outlooks. Sources that pass the metadata checks in Step 2 but fail any execution-level check are rejected and recorded in the rejection log with the failure reason.

\begin{table}[t]
\centering
\caption{Execution-level validity checks in Step 3.}
\label{tab:execution_validation}
\small
\renewcommand{\arraystretch}{1.12}
\begin{tabular}{p{0.045\columnwidth} p{0.895\columnwidth}}
\toprule
\textbf{No.} & \textbf{Validity check} \\
\midrule
1 & \textbf{Period alignment.} The time series value, source release, and question instance refer to the same timestamp, date, period, or update window. \\
2 & \textbf{Minimum history.} The source supports at least \(24\) periods for backtesting and future live evaluation. \\
3 & \textbf{Reference traceability.} Time series analyses in released reports are traceable to concrete public evidence, such as a field, label, value, table, figure, or report statement. \\
4 & \textbf{Information isolation.} The evidence package for period \(t\) uses only information available at or before the query time, excluding future releases, later revisions, and retrospective explanations. \\
5 & \textbf{Objective evaluability.} Time series analyses in released reports can be derived by predefined rules and evaluated without relying on subjective preference. \\
\bottomrule
\end{tabular}
\renewcommand{\arraystretch}{1.0}
\end{table}

\subsubsection{Stage 2: Scenario Generation}

\paragraph{Step 4: Evidence collection.}
Taking each instance skeleton validated in Step~3 as input, the fourth step assembles the evidence state that an agent will see at evaluation time, which also serves as the traceable substrate for deriving ground truth. We preserve the raw source downloads and extract an evidence state comprising two parts: the cumulative time series analysis reports and the cumulative time series table. For period \(t_k\), the report set contains all reports from the earliest available period through \(t_{k-1}\), while the time series table contains all rows from the earliest available period through \(t_k\). The output is therefore a time-isolated evaluation state: the agent receives the cumulative table and the prior-period source documents, while the target-period report is withheld for ground-truth construction. Each scenario yields at least 24 such per-period states, one per instance.

The evidence is accepted only if it passes two checks: \textit{\textbf{(1)}} \textbf{Completeness}, ensuring sufficient coverage, temporal continuity, and plausible numeric scale; and \textit{\textbf{(2)}} \textbf{Column hygiene}, keeping only genuine raw measurements while removing duplicate, constant, textual, rescaled, or already-derived fields unless explicitly allowed.

\paragraph{Step 5: Question generation.}
Using the evidence from Step~4 and the assigned difficulty tier, Step~5 defines a scenario-level evaluation contract: a fixed question template, answer schema, and scoring rule shared across all periods.
Concretely, we use an LLM (Claude Opus 4.6) to match each question intent in Table~\ref{tab:question_taxonomy} against the time series analysis reports and extract the corresponding analysis paragraphs, which the next step transforms into the answer schema and scoring rule listed in Table~\ref{tab:scoring_assignment}.

In the meantime, the question coverage in difficulty tier is enforced: Easy uses at least one A-group tag; Medium uses at least three tags across A/B/C/D plus a universal summary; and Hard uses at least three tags across E/F/D plus a summary and traceable evidence. Before admission, each contract must pass three checks: (1) recurrence across sampled and full periods, (2) synchronization among tags, schema fields, and scoring rules, and (3) artifact grounding against the cumulative table, instance folder, and answer schema. Contracts that fail any check are rejected.

\begin{table}[htbp]
\centering
\caption{Question taxonomy in step 5: 27 fine-grained question types in six main categories (A--F).}
\label{tab:question_taxonomy}
\small
\renewcommand{\arraystretch}{1.18}
\resizebox{\columnwidth}{!}{
\begin{tabular}{p{0.045\columnwidth} p{0.895\columnwidth}}
\toprule
\textbf{ID} & \textbf{Question intent} \\
\midrule
\multicolumn{2}{p{0.94\columnwidth}}{\textit{A --- State identification}} \\
A1 & The current categorical state of the target. \\
A2 & The latest reported value. \\
A3 & The direction of the latest reading (up / down / flat). \\
\midrule
\multicolumn{2}{p{0.94\columnwidth}}{\textit{B --- Change detection}} \\
B1 & The change from the previous period. \\
B2 & The direction of the period-over-period change. \\
B3 & The magnitude of the recent multi-period trend. \\
B4 & How many consecutive periods the trend has lasted. \\
B5 & The deviation from a declared baseline (e.g., five-year average). \\
B6 & Whether this period is a record extremum within a time window. \\
B7 & The ordinal rank of this period within its history. \\
\midrule
\multicolumn{2}{p{0.94\columnwidth}}{\textit{C --- Composition identification}} \\
C1 & The set of leading contributors (top-\(N\)). \\
C2 & The single dominant contributor. \\
C3 & The breakdown of the whole into components with their values or shares. \\
C4 & The assignment of entities into named buckets. \\
\midrule
\multicolumn{2}{p{0.94\columnwidth}}{\textit{D --- Event detection}} \\
D1 & The anomalies or unusual values flagged this period. \\
D2 & The classification or enumeration of discrete events. \\
\midrule
\multicolumn{2}{p{0.94\columnwidth}}{\textit{E --- Attribution}} \\
E1 & The single cause attributed to the change. \\
E2 & The primary driver among several. \\
E3 & The secondary drivers. \\
E4 & The set of drivers, each paired with traceable evidence. \\
E5 & The stakeholders affected by the change. \\
E6 & The issuer's forward guidance or committed response. \\
E7 & A comparison with a named earlier period or season. \\
\midrule
\multicolumn{2}{p{0.94\columnwidth}}{\textit{F --- Outlook}} \\
F1 & The forward risks, paired with traceable evidence. \\
F2 & The direction, regime, or tone of the outlook. \\
F3 & The outlook with its confidence and horizon. \\
F4 & A conditional outlook (``if X, then Y''). \\
\bottomrule
\end{tabular}}
\renewcommand{\arraystretch}{1.0}
\label{tab:question_taxonomy }
\end{table}

\begin{table}[htbp]
\centering
\caption{Scoring-rule assignment by question type in step 5. Rule details are defined in Table~\ref{tab:evaluation_primitives}.}
\label{tab:scoring_assignment}
\small
\renewcommand{\arraystretch}{1.18}
\begin{tabular}{@{}p{0.31\columnwidth} p{0.63\columnwidth}@{}}
\toprule
\textbf{Scoring rule} & \textbf{Question types} \\
\midrule
\multicolumn{2}{@{}l}{\textit{Atomic rules}} \\
Label match       & A1, A3, B2, B6, C2, D2, E1, E2, E5, F2 \\
Numeric check     & A2, B1, B3, B4, B5, B7 \\
Set overlap       & C1, C4, D1, D2, E3, E5 \\
LLM judgment      & E1, period summary \\
\midrule
\multicolumn{2}{@{}l}{\textit{Composition rules}} \\
Joint scoring     & E6, E7, F3, F4 \\
Item-wise scoring & C3, E4, F1 \\
\bottomrule
\end{tabular}
\renewcommand{\arraystretch}{1.0}
\end{table}

\paragraph{Step 6: Ground-truth extraction.}
Step~6 produces a reproducible, leak-free canonical answer. For tasks whose answers can be derived directly from the time series, such as state identification or change detection, we use an LLM (Claude Opus 4.6) to generate executable code (e.g., a rule-based state identification script or a change-rate calculation) that yields verifiable answers. For tasks whose answers must be inferred jointly from textual context and the time series, we also use the LLM to transform the evidence extracted in Step~5 into the structured schema. This procedure constructs the ground truth from the given questions
under the fixed contract, yielding a verifiable per-period task.

The target-period release is used only for ground-truth construction and verification, never as agent-visible evidence. Thus, each field is either rule-reproducible from the cumulative table or explicitly grounded in the withheld source. Finally, the prompt is rendered from the same scenario contract, specifying the task, inputs, rules, requested fields, and JSON schema.

\subsubsection{Stage 3: Quality Control}

\paragraph{Step 7: Grounding audit.}
Step~7 verifies that the question, ground-truth, evidence, data, and source documents are mutually consistent and solvable under the intended information boundary. The audit applies eight gates covering schema, rules, data, evidence, and cross-instance consistency (Table~\ref{tab:grounding_audit_gates_full}); a scenario passes only when all gates have zero-failed outcomes.

\begin{table}[htbp]
\begin{center}
\captionof{table}{The eight grounding-audit gates in Step 7.}
\label{tab:grounding_audit_gates_full}
\small
\renewcommand{\arraystretch}{1.18}
\resizebox{\columnwidth}{!}{
\begin{tabular}{p{0.045\columnwidth} p{0.895\columnwidth}}
\toprule
\textbf{Gate} & \textbf{Grounding-audit check} \\
\midrule
1 & \textbf{Schema conformance.} Ground truth validates against the answer schema; its top-level keys equal the required set, and enum values stay within the declared options. \\
2 & \textbf{File completeness.} Each instance carries its question, ground truth, evidence, time series, and schema files. \\
3 & \textbf{Question--artifact consistency.} Columns, files, and period labels named in the question exist and match the data; output-example keys and enum options equal the schema. \\
4 & \textbf{Rule reproducibility.} The deterministic derivation rule recomputes the stored ground truth (report-grounded fields exempt). \\
5 & \textbf{Time series health.} The cumulative table is strictly ascending in time. \\
6 & \textbf{Evidence--data consistency.} Numeric and key--value references in the evidence resolve to table cells, and ground-truth enum strings appear in the evidence. \\
7 & \textbf{Quote grounding.} Every verbatim quote in the supporting evidence is a substring of the cited source document. \\
8 & \textbf{Cross-instance health.} Instance count matches the scenario, ground-truth fields vary across periods, and each table's last row matches its folder period. \\
\bottomrule
\end{tabular}}
\renewcommand{\arraystretch}{1.0}
\end{center}
\end{table}

\paragraph{Step 8: Human check.}
The final step adjudicates audit warnings and catches design violations that are
difficult to formalize. Human reviewers decide whether each warning is by design,
re-check artifact quality against the full gate of Table~\ref{tab:artifact_quality_gate},
and remove leakage references. By-design warnings may be released, such as hard-tier
evidence warnings, whitelisted constants, or benign period-formatting differences,
whereas scenarios are blocked or quarantined if they contain target-period leakage,
revision-framing leakage, column-hygiene defects, or instance-count defects. After the
human check, accepted scenarios are promoted into the benchmark, portfolio statistics
are recomputed, and the termination condition
\(\texttt{total} \geq 60 \wedge \texttt{domains} \geq 5 \wedge \min(\texttt{tier}) \geq 10\)
is re-checked; if it is not met, the loop returns to Step~1 for the next domain pass.

\begin{table}[htbp]
\begin{center}
\captionof{table}{The nine-section artifact quality gate in Step 8.}
\label{tab:artifact_quality_gate}
\small
\renewcommand{\arraystretch}{1.18}
\resizebox{\columnwidth}{!}{
\begin{tabular}{p{0.045\columnwidth} p{0.895\columnwidth}}
\toprule
\textbf{No.} & \textbf{Artifact quality check} \\
\midrule
1 & \textbf{Time series completeness.} Core columns are populated, the date axis is continuous, and there are at least 24 rows (cf.\ Step 4). \\
2 & \textbf{Column hygiene.} No duplicate date parts, rescaled duplicates, constant columns, embedded narrative, or agent-derivable columns. \\
3 & \textbf{GT leakage.} No column's latest value, value set, or sign restates the ground-truth answer (Easy categorical state exempted). \\
4 & \textbf{Difficulty--task alignment.} Easy reads a latest-row state; Medium summarises across time from a brief; Hard reasons causally from a report, with no analytical conclusions baked into the table. \\
5 & \textbf{Question non-hallucination.} Referenced fields and windows exist; the question recurs across at least three periods and instantiates over all periods; difficulty matches the tier. \\
6 & \textbf{Schema--template consistency.} Tier and domain match the directory, Easy options equal the schema enum, and declared columns equal the table header. \\
7 & \textbf{Directory compliance.} Required scenario, environment, script, source, and per-instance files are present, with at least 24 instances. \\
8 & \textbf{GT traceability.} The latest period's evidence cites concrete source locations, and no field is inferable from the table alone (cf.\ Step 6). \\
9 & \textbf{Information isolation.} No table rows or evidence citations postdate the instance period (V7). \\
\bottomrule
\end{tabular}}
\renewcommand{\arraystretch}{1.0}
\end{center}
\end{table}

\subsubsection{Human Annotations}
\label{app:sec_human_annotations}

We recruit three human reviewers to evaluate the quality of TimeSage-EV. All reviewers are PhD students in Computer Science with at least full professional proficiency in English, and all consent to the use of their annotations for benchmark construction and analysis. The reviewers examine scenario specifications, period-specific questions, structured ground truth, and supporting evidence. The review focuses on whether each question is clear and answerable under the cutoff, whether the ground-truth fields are schema-conformant and correct, and whether each answer is grounded in the cumulative time series or cutoff-valid source documents. Disagreements are resolved through discussion, and a scenario is released only after both automatic checks and human review pass. The following are the detailed instructions to guide the reviewers:

\begin{promptbox}
Thank you for helping verify the quality of TimeSage-EV, a benchmark for recurring time series analysis under a temporal cutoff. Please follow the instructions below.

For each scenario you will receive: a scenario specification, period-specific questions, an answer schema, the ground truth (ground_truth.json) with its evidence (gt_evidence.md), the cumulative timeseries.csv, and any prior-period source documents.

Instructions
1. Review without help from other people, answer-generation tools, or search engines beyond the provided documents. There is no time limit.
2. For each scenario, check:
   (a) Clarity: Is the question clear, recurring, and answerable for the target period?
   (b) Cutoff validity: Can the answer be obtained without the target-period source or any future evidence?
   (c) Schema conformance: Do all ground-truth fields match the schema and allowed values?
   (d) Correctness: Are the structured answers correct for the target period?
   (e) Grounding: Is every field supported by the time series or cutoff-valid documents?
   (f) Variation: Does the scenario vary across periods enough for recurring evaluation?
3. If a scenario is unclear, incorrect, or ambiguous, flag it with the affected period, field, and a short explanation.
4. Return one of: PASS, PASS_WITH_MINOR_FIXES, or FAIL.

Thank you for your contributions!
\end{promptbox}


\newcommand{\tiereasy}{\textsc{e}}
\newcommand{\tiermid}{\textsc{m}}
\newcommand{\tierhard}{\textsc{h}}

\definecolor{groupgray}{gray}{0.90}
\newcounter{groupseen}
\newcommand{\grouprow}[1]{%
  \multicolumn{4}{c}{\cellcolor{groupgray}\textbf{#1}}\\*
  \noalign{\stepcounter{groupseen}}}

\setlength{\tabcolsep}{2pt}
\newlength{\tcolT}\setlength{\tcolT}{0.62cm}
\newlength{\tcolS}\setlength{\tcolS}{2.55cm}
\newlength{\tcolC}\setlength{\tcolC}{0.78cm}
\newlength{\tcolU}%
\setlength{\tcolU}{\dimexpr\columnwidth-\tcolT-\tcolS-\tcolC-6\tabcolsep\relax}
\newcolumntype{T}{>{\centering\arraybackslash\scriptsize}p{\tcolT}}
\newcolumntype{S}{>{\RaggedRight\arraybackslash\ttfamily\scriptsize}p{\tcolS}}
\newcolumntype{C}{>{\RaggedRight\arraybackslash\scriptsize}p{\tcolC}}
\newcolumntype{U}{>{\RaggedRight\arraybackslash\ttfamily\scriptsize}p{\tcolU}}

\renewcommand{\arraystretch}{1.0}

\begin{table}[htbp]
\centering

\begingroup
\sloppy\setlength{\emergencystretch}{1.5em}\scriptsize

\caption{Public data sources for all 60 TimeSage scenarios, grouped by domain.
Tier: \tiereasy~= easy, \tiermid~= middle, \tierhard~= hard. Scenario names are
shortened to the shortest unique prefix.}
\label{tab:data_sources}

\setcounter{groupseen}{0}%
\begin{tabular}{@{}T S C U@{}}
\toprule
\textbf{T} & \textbf{Scenario} & \textbf{Cad.} & \textbf{Source} \\
\midrule
\grouprow{Agriculture}
\tiereasy & usda\_\allowbreak cattle & mon. & esmis.\allowbreak nal.\allowbreak usda.\allowbreak gov \\
\tiereasy & usda\_\allowbreak chicken & mon. & esmis.\allowbreak nal.\allowbreak usda.\allowbreak gov \\
\tiereasy & usda\_\allowbreak cold\_\allowbreak storage\_\allowbreak butter & mon. & esmis.\allowbreak nal.\allowbreak usda.\allowbreak gov \\
\tiereasy & usda\_\allowbreak cold\_\allowbreak storage\_\allowbreak pork & mon. & esmis.\allowbreak nal.\allowbreak usda.\allowbreak gov \\
\tiermid & usda\_\allowbreak crop & wkly. & esmis.\allowbreak nal.\allowbreak usda.\allowbreak gov \\
\tiermid & usda\_\allowbreak drought & wkly. & droughtm\allowbreak onitor.\allowbreak unl.\allowbreak edu \\
\tiermid & usda\_\allowbreak wasde & mon. & usda.\allowbreak gov \\
\tierhard & fao\_\allowbreak food & mon. & fao.\allowbreak org \\
\tierhard & usda\_\allowbreak fas & mon. & apps.\allowbreak fas.\allowbreak usda.\allowbreak gov \\
\tierhard & usda\_\allowbreak us & mon. & ams.\allowbreak usda.\allowbreak gov;\allowbreak\ fred.\allowbreak stlouisfed.\allowbreak org \\
\grouprow{Climate}
\tiereasy & noaa\_\allowbreak cpc\_\allowbreak ao & mon. & cpc.\allowbreak ncep.\allowbreak noaa.\allowbreak gov \\
\tiereasy & noaa\_\allowbreak cpc\_\allowbreak mjo & daily & psl.\allowbreak noaa.\allowbreak gov \\
\tiereasy & noaa\_\allowbreak cpc\_\allowbreak oni & mon. & cpc.\allowbreak ncep.\allowbreak noaa.\allowbreak gov \\
\tiereasy & noaa\_\allowbreak cpc\_\allowbreak pna & mon. & cpc.\allowbreak ncep.\allowbreak noaa.\allowbreak gov \\
\tiermid & c3s\_\allowbreak copernicus & mon. & climate.\allowbreak copernicus.\allowbreak eu \\
\tiermid & noaa\_\allowbreak ncei\_\allowbreak global & mon. & ncei.\allowbreak noaa.\allowbreak gov \\
\tiermid & noaa\_\allowbreak ncei\_\allowbreak us & mon. & ncei.\allowbreak noaa.\allowbreak gov \\
\tierhard & iri\_\allowbreak enso & mon. & iri.\allowbreak columbia.\allowbreak edu \\
\tierhard & noaa\_\allowbreak cpc\_\allowbreak enso & mon. & cpc.\allowbreak ncep.\allowbreak noaa.\allowbreak gov \\
\tierhard & noaa\_\allowbreak cpc\_\allowbreak monthly & mon. & cpc.\allowbreak ncep.\allowbreak noaa.\allowbreak gov \\
\grouprow{Energy}
\tiereasy & eia\_\allowbreak refinery & wkly. & eia.\allowbreak gov \\
\tiereasy & eia\_\allowbreak us\_\allowbreak commercial & wkly. & eia.\allowbreak gov \\
\tiereasy & eia\_\allowbreak wti & daily & fred.\allowbreak stlouisfed.\allowbreak org \\
\tiermid & eia\_\allowbreak natural\_\allowbreak gas\_\allowbreak monthly & mon. & eia.\allowbreak gov \\
\tiermid & eia\_\allowbreak us\_\allowbreak petroleum & mon. & eia.\allowbreak gov \\
\tiermid & eia\_\allowbreak wpsr & wkly. & eia.\allowbreak gov \\
\tierhard & eia\_\allowbreak monthly & mon. & eia.\allowbreak gov \\
\tierhard & eia\_\allowbreak natural\_\allowbreak gas\_\allowbreak weekly & wkly. & eia.\allowbreak gov \\
\tierhard & iea\_\allowbreak oil & mon. & iea.\allowbreak org \\
\tierhard & ief\_\allowbreak comparative & mon. & ief.\allowbreak org \\
\grouprow{Finance}
\tiereasy & cboe\_\allowbreak vix & daily & fred.\allowbreak stlouisfed.\allowbreak org \\
\tiereasy & fred\_\allowbreak sahm & mon. & fred.\allowbreak stlouisfed.\allowbreak org \\
\tiereasy & fred\_\allowbreak us & daily & fred.\allowbreak stlouisfed.\allowbreak org \\
\tiermid & dallas\_\allowbreak fed & mon. & dallasfed.\allowbreak org \\
\tiermid & fred\_\allowbreak consumer & mon. & fred.\allowbreak stlouisfed.\allowbreak org \\
\tiermid & philly\_\allowbreak fed & mon. & philadel\allowbreak phiafed.\allowbreak org \\
\tierhard & bea\_\allowbreak pce & mon. & fred.\allowbreak stlouisfed.\allowbreak org \\
\tierhard & boe\_\allowbreak monetary & 8/yr & bankofen\allowbreak gland.\allowbreak co.\allowbreak uk \\
\tierhard & ecb\_\allowbreak monetary & 8/yr & ecb.\allowbreak europa.\allowbreak eu \\
\tierhard & frb\_\allowbreak us & mon. & fred.\allowbreak stlouisfed.\allowbreak org \\
\grouprow{Healthcare}
\tiereasy & cdc\_\allowbreak nwss & wkly. & data.\allowbreak cdc.\allowbreak gov \\
\tiereasy & cdc\_\allowbreak us & wkly. & data.\allowbreak cdc.\allowbreak gov \\
\tiereasy & hhs\_\allowbreak us & wkly. & data.\allowbreak cdc.\allowbreak gov \\
\tiermid & bls\_\allowbreak jolts & mon. & fred.\allowbreak stlouisfed.\allowbreak org \\
\tiermid & bls\_\allowbreak us & mon. & fred.\allowbreak stlouisfed.\allowbreak org \\
\tiermid & ecdc\_\allowbreak cdtr & wkly. & ecdc.\allowbreak europa.\allowbreak eu \\
\tiermid & paho\_\allowbreak measles & biwk. & paho.\allowbreak org \\
\tierhard & bls\_\allowbreak medical & mon. & fred.\allowbreak stlouisfed.\allowbreak org \\
\tierhard & cdc\_\allowbreak resp & wkly. & data.\allowbreak cdc.\allowbreak gov \\
\tierhard & who\_\allowbreak cholera & mon. & who.\allowbreak int \\
\grouprow{Transportation}
\tiereasy & bls\_\allowbreak diesel & wkly. & fred.\allowbreak stlouisfed.\allowbreak org \\
\tiereasy & bts\_\allowbreak airline & mon. & fred.\allowbreak stlouisfed.\allowbreak org \\
\tiereasy & tsa\_\allowbreak checkpoint & daily & tsa.\allowbreak gov \\
\tiermid & ata\_\allowbreak truck & mon. & fred.\allowbreak stlouisfed.\allowbreak org \\
\tiermid & bls\_\allowbreak ppi & mon. & fred.\allowbreak stlouisfed.\allowbreak org \\
\tiermid & eurocontrol\_\allowbreak monthly & mon. & eurocont\allowbreak rol.\allowbreak int \\
\tiermid & portla\_\allowbreak monthly & mon. & portoflo\allowbreak sangeles.\allowbreak org \\
\tierhard & cass\_\allowbreak freight & mon. & cassinfo.\allowbreak com \\
\tierhard & freightos\_\allowbreak weekly & wkly. & freightos.\allowbreak com \\
\tierhard & iata\_\allowbreak air & mon. & iata.\allowbreak org \\
\bottomrule
\end{tabular}
\endgroup
\end{table}

\begin{figure}[thbp]
    \centering
    \includegraphics[width=\linewidth]{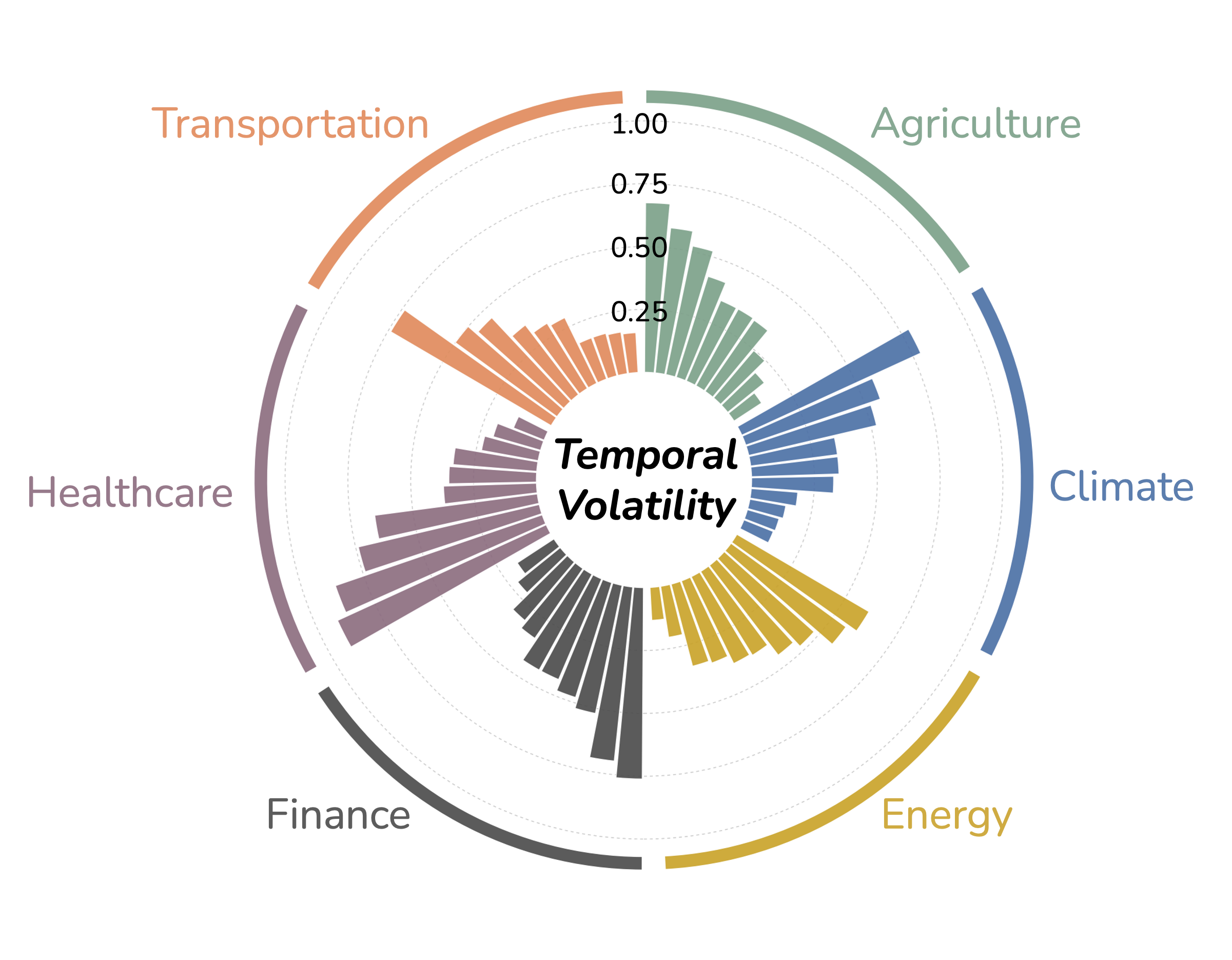}
    \caption{Per-Scenario Answer Volatility Across Domains.}
    \label{fig:temporal_volatility}
\end{figure}

\subsubsection{Released Sources}
\label{appendix_data_sources}

After the construction loop, Table~\ref{tab:data_sources} lists the sources that pass metadata filtering, execution validation, ground-truth extraction, automatic audits, and human review. These sources form the released TimeSage-EV corpus, which contains \(60\) scenarios across six domains, including agriculture, climate, energy, finance, healthcare, and transportation, and three tiers, with \(20\) scenarios per tier. 

We quantify how much each environment evolves over time using the per-period answer transition rate (Figure~\ref{fig:temporal_volatility}), which measures how often a scenario's structured answer fields change between consecutive periods. Across all 60 scenarios the rate spans 0.13 to 0.91, confirming that temporal volatility is not uniform but a scenario-level property the benchmark deliberately varies.

\begin{figure}[thbp]
    \centering
    \includegraphics[width=\linewidth]{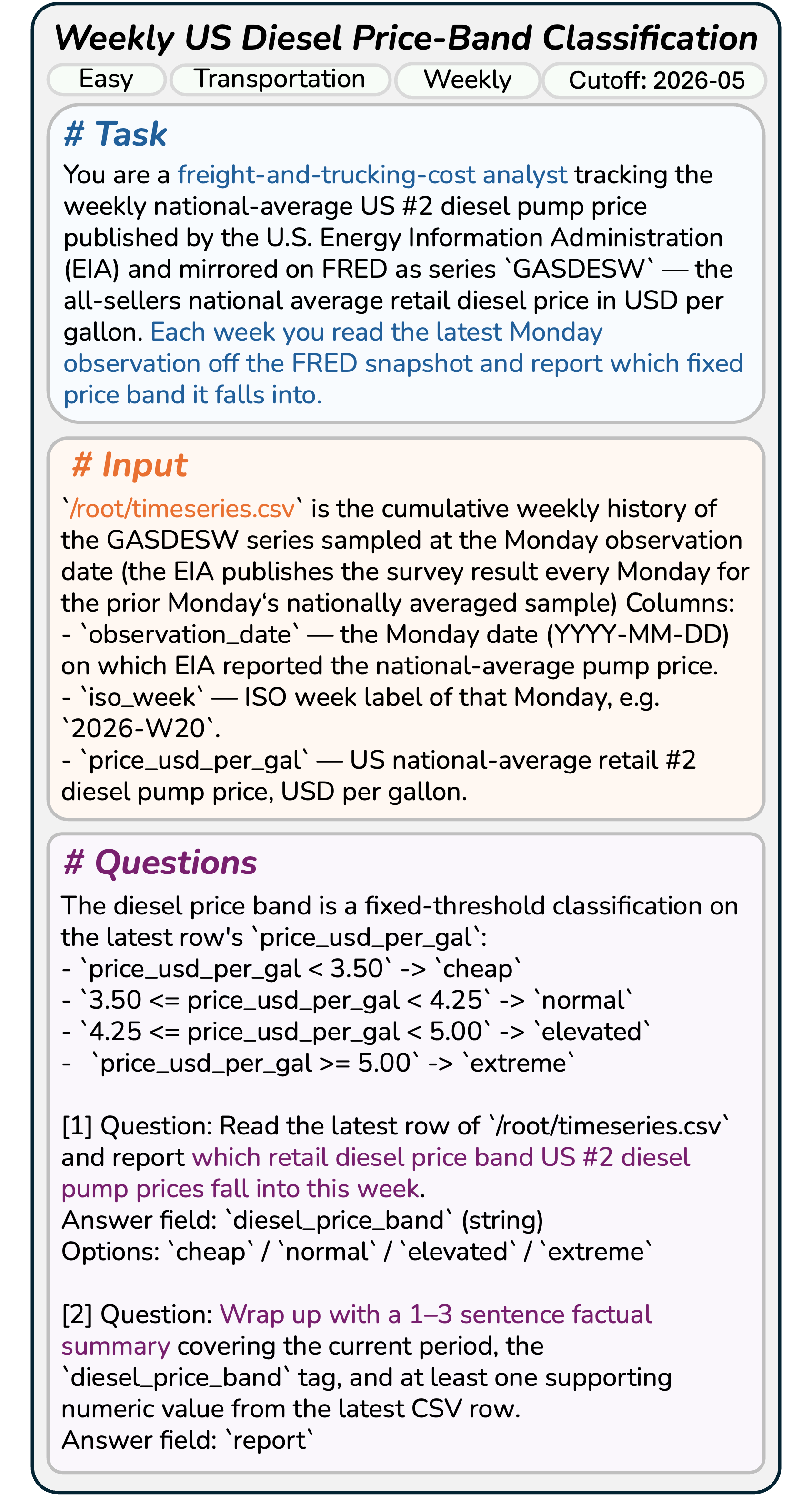}
    \caption{Easy scenario: Weekly US Diesel Price-Band Classification.}
    \label{fig:easy_scenario}
\end{figure}

\begin{figure}[htbp]
    \centering
    \includegraphics[width=\linewidth]{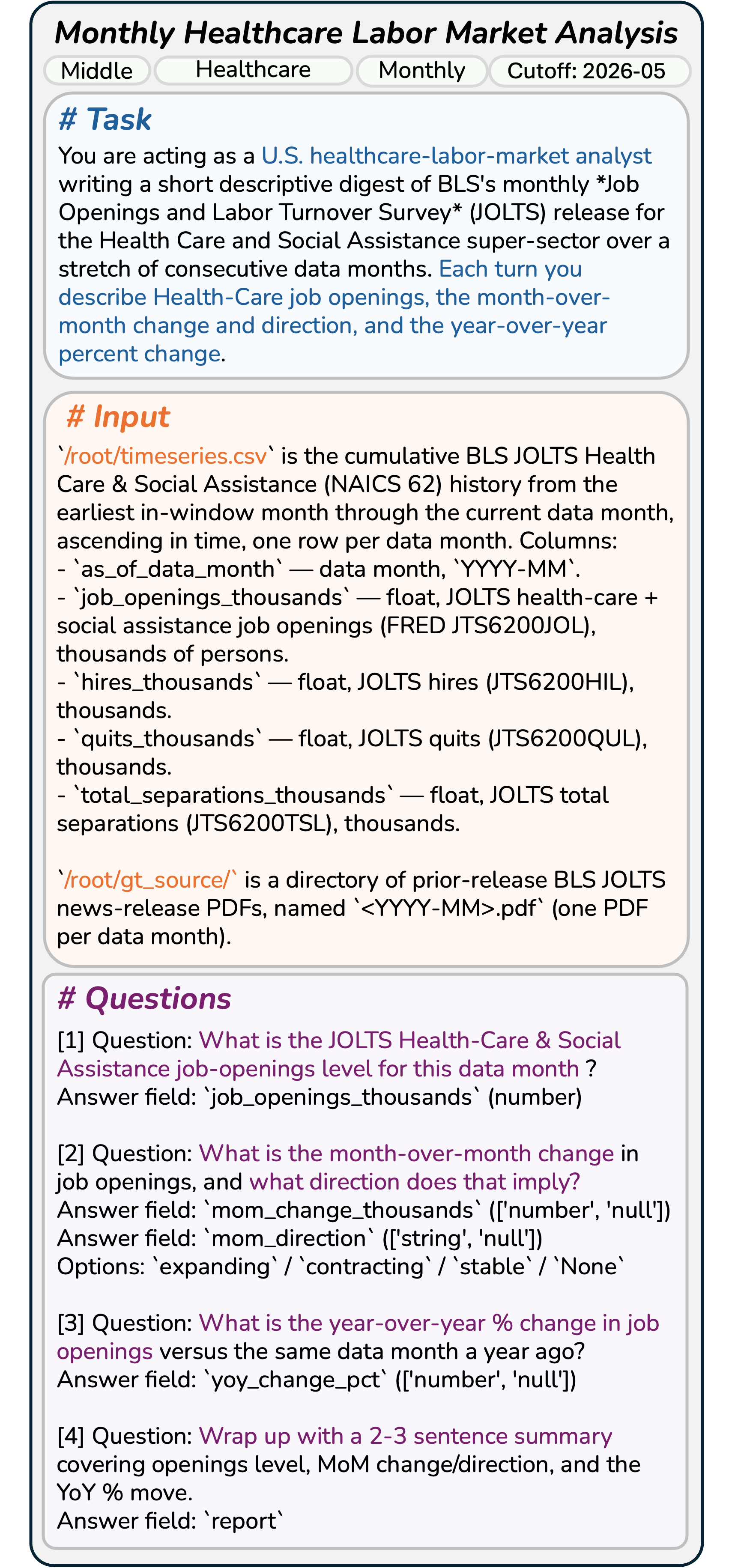}
    \caption{Medium scenario: Monthly Healthcare Labor Market Analysis.}
    \label{fig:middle_scenario}
\end{figure}

\begin{figure}[thbp]
    \centering
    \includegraphics[width=\linewidth]{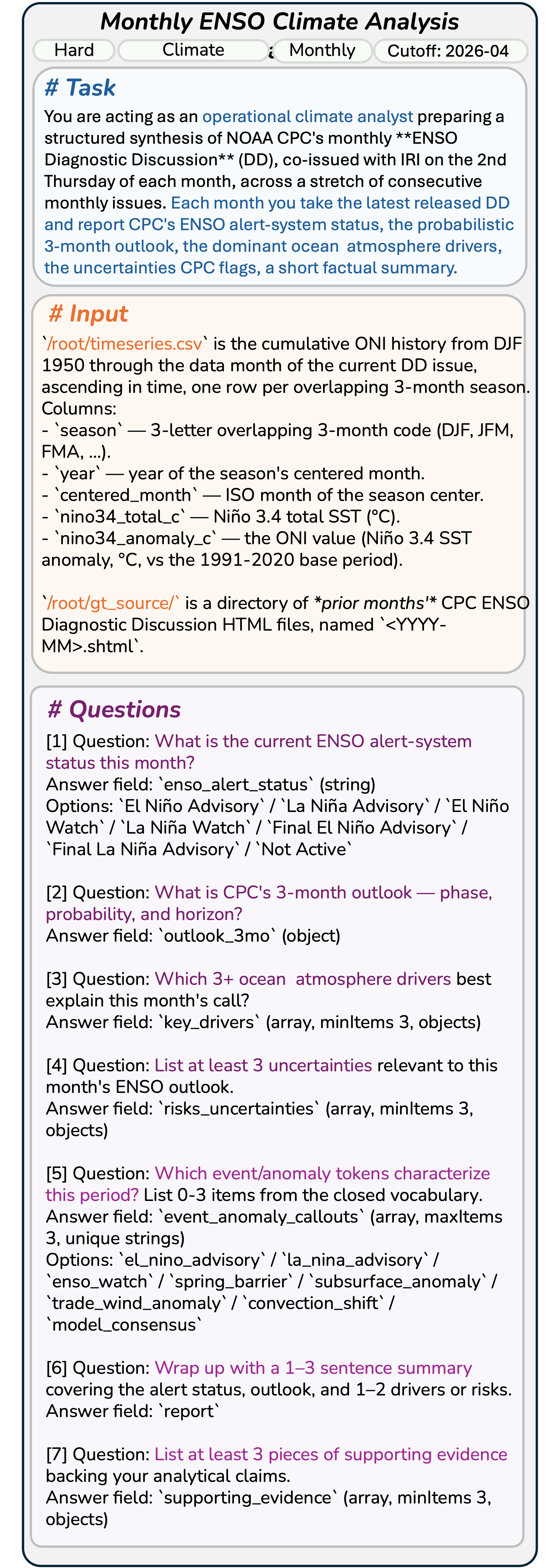}
    \caption{Hard scenario: Monthly ENSO Climate Analysis.}
    \label{fig:hard_scenario}
\end{figure}

\subsubsection{Scenario Examples}
\label{app:sec_scenario_examples}

To illustrate the range of tasks covered by our benchmark, Figures~\ref{fig:easy_scenario}–\ref{fig:hard_scenario} present three representative scenarios spanning the Easy, Medium, and Hard difficulty tiers. Each scenario casts the model as a domain analyst who must read a cumulative time-series snapshot and analysis reports and answer a structured set of questions about the latest reporting period. 

The Easy scenario (Weekly US Diesel Price-Band Classification) in Figure~\ref{fig:easy_scenario} asks the model to read the most recent row of an EIA/FRED diesel price series and assign it to one of four fixed price bands, followed by a brief  summary. This is a task that relies on data lookup and threshold logic. 

The Medium scenario (Monthly Healthcare Labor Market Analysis) in Figure~\ref{fig:middle_scenario} draws on BLS JOLTS data for the Health Care and Social Assistance sector and additionally requires the model to compute month-over-month and year-over-year changes, infer their direction, and synthesize a short digest, thereby introducing multi-step quantitative reasoning over the time series. 


The Hard scenario (Monthly ENSO Climate Analysis), shown in Figure~\ref{fig:hard_scenario}, is the most open-ended. Working from NOAA CPC's monthly ENSO Diagnostic Discussion Reports alongside the Niño 3.4 ONI time series history, the model must report six things: the current alert-system status, a three-month outlook, the dominant ocean–atmosphere drivers, any flagged uncertainties, event tokens drawn from a closed vocabulary, and a grounded summary with supporting evidence.

\subsection{Evaluation Methodology}
\label{appendix_evaluation}

We evaluate each agent run along four independent axes to disentangle answer correctness from report quality: structured-answer accuracy on the one hand, and report coverage, faithfulness, and writing quality on the other. This section first defines the evaluation unit in Section~\ref{appendix_eval_unit}, then detail the evaluation method for each axis in Section~\ref{appendix_eval_axes}.

\subsubsection{Evaluation Unit}
\label{appendix_eval_unit}


The atomic unit of evaluation is a \emph{(scenario, period)} instance. For each instance, the agent produces two artifacts: a structured answer that fills the scenario answer schema and a free-form report that summarizes the answer. The answer schema decomposes each period time series analysis into a set of keypoints corresponding to the elementary facts that a competent analysis must establish. The report serves as an analytical summary, integrating these keypoints and their evidence into human-readable text. We evaluate the two artifacts along four axes, each scored on a $0$--$100$ scale: keypoint accuracy (\textsc{Acc}), keypoint coverage (\textsc{Cov}), report faithfulness (\textsc{Faith}), and report quality (\textsc{Qual}). Averaging them gives the overall instance score:

\begin{equation*}
\textsc{Agg} =
\operatorname{mean}\bigl(\textsc{Acc},\textsc{Cov},\textsc{Faith},\textsc{Qual}\bigr).
\end{equation*}

\subsubsection{The Four Evaluation Axes}
\label{appendix_eval_axes}

\paragraph{Keypoint accuracy.}
This axis measures how well the structured answer matches the ground truth (GT). Each keypoint is scored by comparing the agent-provided value with the GT value according to the answer-type-specific rules in Table~\ref{tab:evaluation_primitives}. The match score reflects the degree of agreement between the two values, and the final score is the average across keypoints, rescaled to $0$--$100$.

\begin{table}[htbp]
\centering
\caption{Primitive match rules. Each rule compares the agent output \(a\) with the ground-truth value \(b\), or output set \(A\) with ground-truth set \(B\).}
\label{tab:evaluation_primitives}
\small
\setlength{\tabcolsep}{4pt}
\renewcommand{\arraystretch}{1.18}
\begin{tabularx}{\columnwidth}{@{}p{0.25\columnwidth} X@{}}
\toprule
\textbf{Rule} & \textbf{Match score} \\
\midrule
Label match &
\(1\) if \(a=b\), else \(0\). \\
Boolean match &
\(1\) if \(\operatorname{bool}(a)=\operatorname{bool}(b)\), else \(0\). \\
Text match &
\(1\) if stripped strings are equal, else \(0\). \\
Numeric check &
\(1\) if \(|a-b|\leq\tau_{\text{abs}}\), or \(|a-b|/|b|\leq\tau_{\text{rel}}\) when relative tolerance is set and \(b\neq0\); else \(0\). With no tolerance, exact equality is required. \\
Set overlap &
Dice/F1 overlap \(2|A\cap B|/(|A|+|B|)\); two empty sets score \(1\), and a single empty set scores \(0\). \\
\bottomrule
\end{tabularx}
\renewcommand{\arraystretch}{1.0}
\end{table}

\paragraph{Keypoint coverage.}
This axis measures the report's completeness: whether it communicates the content of each required keypoint. For each keypoint, the judge assigns one of three labels: \textsc{Yes}, \textsc{Partial} or \textsc{No}, which are scored as $1.0$, $0.5$ and $0.0$, respectively. The axis score is obtained by averaging these keypoint-level scores within the instance and rescaling the result to the range $0$--$100$. During the evaluation, the easy and medium tiers contain no reasoning questions. Therefore, a keypoint anchored to a ground-truth scalar is marked covered without an LLM call. In contrast, the hard-tier uses the LLM-based evaluation path, since a verbatim scalar match alone does not guarantee that the underlying concept has been conveyed. The coverage judge prompt is given in Appendix~\ref{appendix:evaluation_prompts}.

\paragraph{Report faithfulness.}
This axis measures factual groundedness: the fraction of the report's
atomic claims supported by the sources. The judge first decomposes the report into a
small set of atomic single-fact claims and then verifies each against a layered
reference: the scenario keypoints (level~1), excerpts of scenario context (level~2),
and general world knowledge (level~3). A claim grounded at any level counts as
supported; claims that contradict a keypoint or are mere speculation count as
unsupported. The axis is the fraction of supported claims, on a scale of $0$--$100$. The judge prompts are presented in the Appendix~\ref{appendix:evaluation_prompts}.

\paragraph{Report quality.}
This axis measures writing quality alone, namely genre conformance, structure, and readability. The judge scores the report against the scenario tier's rubric, which lists the genre, the signals a good report should contain, those it should avoid, and an exemplar; it marks each required signal as fully hit, partially hit, or missed, and flags any forbidden signal. The axis rewards hits, half-credits partial hits,  deducts violations, and normalizes by the number of required signals, on a scale $0$ -- -$100$. Factual errors are not penalized here since the other axes handle them. The quality judge prompt is given in the Appendix~\ref{appendix:evaluation_prompts}.

\subsubsection{Evaluation Examples}

\begin{figure}[htbp]
    \centering
    \includegraphics[width=\linewidth]{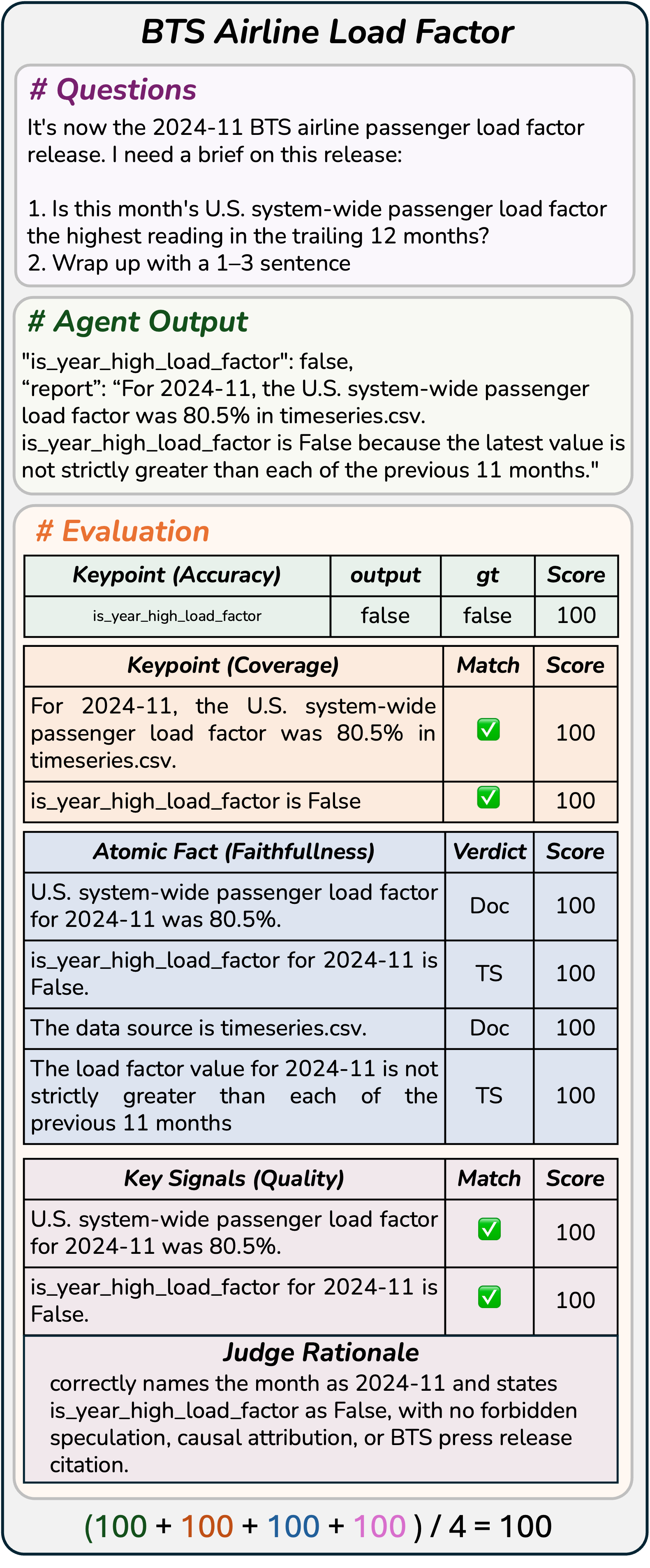}
    \caption{Successful case of agent output.}
    \label{fig:evaluation_right}
\end{figure}

Figure~\ref{fig:evaluation_right} shows a successful run on the BTS Airline 
Load Factor scenario. The agent correctly fills the structured answer 
(is \texttt{\_year \_high \_load \_factor} = false), matching the ground truth and 
producing a perfect keypoint precision. Its report communicates every required 
keypoint—the 80.5\% load factor for 2024-11 and the negative high-load flag—so 
coverage is also full. In faithfulness, each atomic claim is grounded either in 
the keypoints (\textsc{TS}) or the scenario context (\textsc{Doc}), with no 
unsupported speculation. The report additionally hits all required quality 
signals while avoiding forbidden ones, such as causal attribution or fabricated 
press-release citations. All four axes score $100$, giving an aggregate of 
$\textsc{Agg}=100$.

\begin{figure}[htbp]
    \centering
    \includegraphics[width=0.98\linewidth]{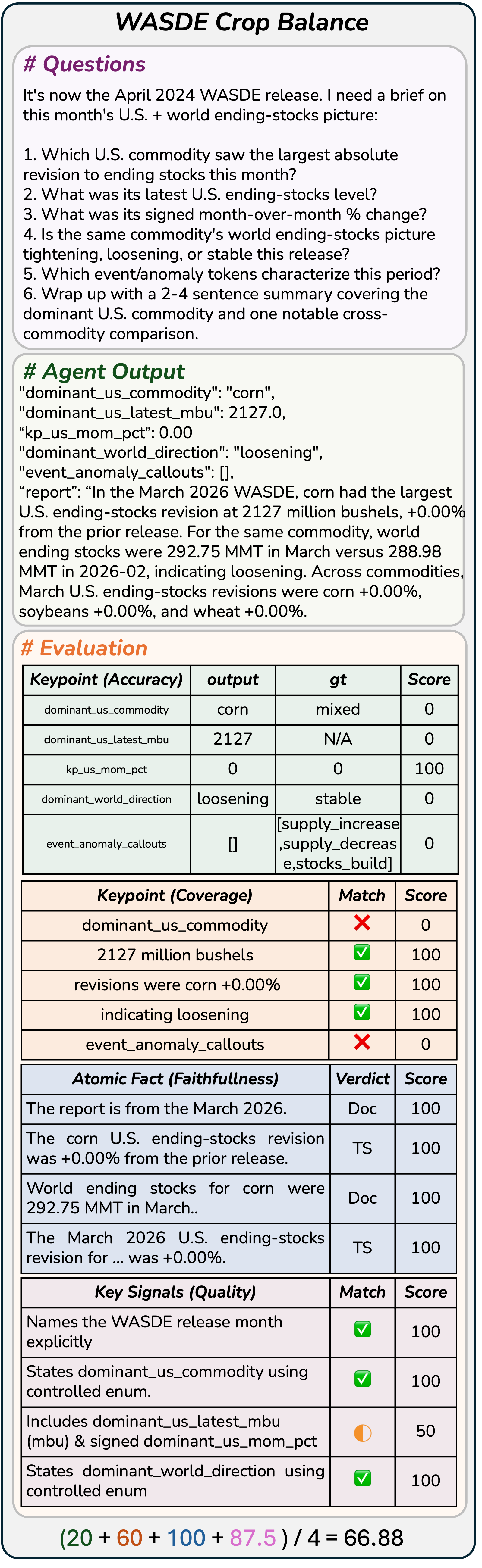}
    \vspace{-10pt}
    \caption{Failed case of agent output.}
    \label{fig:evaluation_failed}
\end{figure}

Figure~\ref{fig:evaluation_failed} shows a failure on the WASDE Crop Balance 
scenario, where the agent over-commits to a single commodity. It answers 
\texttt{corn} for the dominant U.S. commodity (GT: \texttt{mixed}) and reports a 
concrete ending stock level where the ground truth is \texttt{N / A}, so several 
precision keypoints score $0$; only \texttt{kp \_us \_mom \_pct} matches. Coverage is correspondingly partial: the report conveys the level and month-over-month 
keypoints, but omits the event/anomaly callouts and the correct dominant-commodity 
framing. Faithfulness remains high, as the stated claims are individually grounded 
in the keypoints or context. Quality is satisfied but only half-credited on 
the signal requiring both the latest level and the signed monthly change. The 
axis scores ($20$, $60$, $100$, $87.5$) average to $\textsc{Agg}=66.88$, 
illustrating how the four axes jointly separate answer correctness from report 
quality.

\subsection{Prompt Templates}
\label{appendix_prompts}

This appendix reports the prompt templates used by TimeSage-EV.
Scenario-specific prompts are generated from these templates and the scenario specification.
The fully rendered period prompts for all 1,485 period-level instances are released with the benchmark files.

\subsubsection{Agent Prompt}

\paragraph{System and sandbox instruction.}
At each period, the agent receives a task prompt together with the following sandbox instruction.
scenario specific fields such as required answer keys, tier, and optional web-search availability are filled out from the scenario specification.

\begin{promptbox}
You are a markets/data analyst. All files you need are in the CURRENT WORKING DIRECTORY.
The cwd contains:
  * timeseries.csv          -- the historical data series
  * answer_schema.json      -- required JSON shape for your answer
  * gt_source/<prior>.* -- prior-period reports (PDF/HTML/JSON)

Available tools:
  * Python code execution
  * final_answer(answer=<dict>, report=<str>) -- submit your final output.
    You must call this tool exactly once to finish. Do not write the answer to a file.
    Do not emit a final textual reply instead of calling the tool.
  * web_search(query: str, n: int=5) -- audited, time-isolated, enabled only
    for scenarios with audited cutoff-valid retrieval.

Sandbox rules:
  Authorized imports: pandas, numpy, json, csv, re, math, statistics, datetime, pathlib, io, collections, itertools, functools, pypdf.
  Forbidden builtins: open, eval, exec, compile, __import__, globals, locals.
  Forbidden modules include os, sys, subprocess, socket, urllib, requests.

File / IO replacements:
  * Read text:  Path('x.txt').read_text()
  * Read JSON:  json.loads(Path('x.json').read_text())
  * Read CSV:   pd.read_csv('x.csv')
  * Read PDF:   PdfReader('x.pdf').pages[0].extract_text()
  * List dir:   [p.name for p in Path.cwd().iterdir()]

Scenario tier: {tier}
Required answer keys: {required_keys}

Task rules:
  1. Read the CSV and prior-period gt_source files if needed. Do not hallucinate values.
  2. Never cite or read the current period's gt_source. Only prior-period source documents are allowed.
  3. Your final answer dict must contain exactly the required keys. Your report must cite specific CSV cells or prior-period files.
  4. End your run by calling final_answer(answer=..., report=...).
\end{promptbox}

\paragraph{Memory instruction.}
For sequential evaluation, the following memory instruction is appended.

\begin{promptbox}
Memory management is optional in multi-period episodes. If your conversation memory becomes bloated with failed code attempts, long observations, or patterns already learned, you may call request_memory_compaction(reason=...).
The framework will summarize what you have learned across past periods after the current period finishes, then reset memory to that summary before the next period. Do not call this for a single-period task.
\end{promptbox}

\paragraph{Skill prompt.}
When the skill library is enabled, a skill-index block is prepended to each period's task prompt. The block has two forms, depending on whether the library is empty or populated. An agent may save a reusable function with \texttt{induct\_skill} and inspect the full body of a saved skill with \texttt{retrieve\_skill}; saved skills become callable as live tools starting the next period. 

When the library is empty, the following seeding nudge is shown.

\begin{promptbox}
No skills have been saved yet for this sweep. You are the first agent to run.

If you write a small, reusable idiom this period (a schema-specific CSV loader,
a date-to-period converter, a band classifier, a row lookup), save it with
induct_skill so later periods can call it as a tool.

Good candidates: small single-purpose functions returning a clean Python value.
Bad candidates: scenario-specific one-off math, or the literal answer dict.

Induct the skill BEFORE calling final_answer, while the code is fresh.
\end{promptbox}

When there is at least one skill, the reuse-enforcement form is shown instead.

\begin{promptbox}
=== Available skills ===
The following skills are already registered as tools for this period.

INVOKE THESE SKILLS -- do NOT re-implement a listed skill inline. Reuse is
faster, less error-prone, and consistent across periods.

Example call:
  result = {first_skill_name}(...)

Skills:
  - {skill_name} -- {skill_description}
  ...

Use retrieve_skill(name=...) to read a skill's notes and full code before
invoking if its description is not clear enough. If you find a new reusable
idiom this period, save it with induct_skill.
\end{promptbox}

\paragraph{Period question prompt.}
Each period-level question is rendered from the scenario's question template.

\begin{promptbox}
It is now {period_label}. I need an evidence-based analysis for this period's {release_name}.

Use /root/timeseries.csv and any prior-period files in /root/gt_source/.
The current-period source document is not available.

Please answer the following questions:
1. {field_question_1}
2. {field_question_2}
...
N. {field_question_N}

Return a JSON object conforming to /root/answer_schema.json and a concise natural-language report grounded in cutoff-valid evidence.
\end{promptbox}

\subsubsection{Evaluation Prompts}
\label{appendix:evaluation_prompts}

\paragraph{Keypoint coverage judge.}
For report coverage, deterministic literal matching is used when possible. Otherwise, the LLM judge receives the following prompt for each keypoint.

\begin{promptbox}
You are checking whether a report communicates a predefined content unit (ACU). Reply with exactly one of: YES, PARTIAL, NO.

Content unit:
  description: {keypoint_description}
  verification question: {verification_question}

Report:
{agent_report}

Reply with the single word YES, PARTIAL, or NO and nothing else.
\end{promptbox}

\paragraph{Report faithfulness decomposition.}
For report faithfulness, the report is first decomposed into atomic claims.

\begin{promptbox}
Decompose this report into AT MOST 10 atomic, verifiable claims. Each claim must be ONE scalar fact: one number, one named entity, one date, one direction, one classification -- never a compound sentence with multiple facts.

Skip markdown headers, bullet labels, rhetorical sentences, hedges, and definitional preambles. Each claim should be standalone and grounded enough to verify against a data source.

Return a JSON list of strings, no prose around it.

Report:
{agent_report}
\end{promptbox}

\paragraph{Report faithfulness verification.}
Each atomic claim is then verified against cutoff-valid reference sources.

\begin{promptbox}
Verify a single atomic factual claim against reference sources.

Claim:
{atomic_claim}

Reference sources:
{reference_sources}

Pick exactly one verdict from this set:
tier1_supported (keypoint says so),
tier2_supported (context says so),
tier3_general_knowledge (only if the claim is encyclopedic and not contradicting any reference),
contradicts_keypoint (clearly conflicts with a keypoint),
speculation (unsupported by any reference).

Return JSON: {"verdict": "<one>"}
\end{promptbox}

\paragraph{Report quality judge.}
Report quality is scored separately from factual correctness.

\begin{promptbox}
You are scoring a report's WRITING quality only -- its genre conformance, structure, and readability. DO NOT score for factual correctness; that is evaluated separately by a different scorer.

Tier genre:
  {tier_genre}

Required signals:
  - {required_signal_1}
  - {required_signal_2}
  ...

Forbidden signals:
  - {forbidden_signal_1}
  - {forbidden_signal_2}
  ...

Exemplar of a good report:
  {quality_exemplar}

Report to score:
{agent_report}

Return JSON:
{
  "required_hits": ["<signal>", ...],
  "required_partial": ["<signal>", ...],
  "required_misses": ["<signal>", ...],
  "forbidden_violations": ["<signal>", ...],
  "rationale": "<one sentence>"
}
\end{promptbox}

\subsection{Agent Harnesses}
\label{appendix:experiment_setting}

\subsubsection{Baseline Harness: \texttt{smolagent} } 

\paragraph{Agent framework.} All agents are built on \texttt{smolagents}, a
lightweight framework in which an LLM drives a multi-step loop of tool calls,
each step appended to a running memory of prior actions and observations. We use
its code-agent variant, where the model acts by writing executable Python rather
than emitting structured tool-call arguments, which suits the data-manipulation
nature of time series analysis and exposes a single uniform action channel
across all models.

\paragraph{Action surface.} The baseline agent is a \texttt{smolagent} Code
Agent in which Python is the action language: the agent emits code, the sandbox
executes it, and the resulting observation becomes the next turn. Imports are
restricted to a whitelist of analytical libraries, file access is mediated
through path-based readers rather than raw file handles, and all networking
modules are excluded by design, so the only sanctioned external access is the
audited web-search tool. No default tools are added beyond those the protocol
requires: the agent can submit its output, request memory compaction, and issue
web searches.

\paragraph{Memory management.} Memory management uses a hybrid policy: a
compaction is scheduled at the next period boundary whenever the agent requests
that one or the most recent step's input-token count exceed a fixed threshold. When
triggered, the harness issues a single call to the agent's own model to
summarize the CSV layout, working code idioms, schema pitfalls, feedback
patterns, and a one-line recap per past period, then re-seeds the conversation
with that summary.

\paragraph{Internet access.} Our harness blocks all network libraries
(e.g., \texttt{urllib}, \texttt{requests}, \texttt{socket}). To preserve
a broad action space, we expose a web-search tool as the sole network
channel, which returns only historical archived snapshots through a
three-stage pipeline: (1) Candidate discovery:
SearXNG\footnote{\url{https://searxng.org}} resolves the query into a
set of candidate URLs; (2) Archive retrieval:
each URL is queried against Common
Crawl\footnote{\url{https://commoncrawl.org}} with server-side temporal
filtering, so that only captures predating the cutoff are returned;
(3) Fallback: if no qualifying capture
exists, the Wayback Machine\footnote{\url{https://web.archive.org}}
supplies the snapshot closest to the cutoff, which is discarded if dated
after it. These stages guarantee that every retrieved page was
archived strictly before the cutoff date.

\subsubsection{Self-Evolving Harness: TimeSage-1.0}

\paragraph{Why Self-Evolve.} TimeSage-EV evaluates agents on a recurring analytical task whose evidence changes over time, raising a question that single-period evaluation cannot answer: \textit{can an agent improve future analyses by converting feedback and repeated reasoning patterns from earlier periods into reusable procedures?} A standard \texttt{smolagent} harness with tool use and memory carries previous context, but mixes reusable routines with transient observations, failed code attempts, and period-specific facts, so memory alone cannot distinguish a generally useful workflow from merely accumulated history.

\paragraph{A diagnostic harness.} TimeSage-1.0 addresses this question by keeping the agent interface, model configuration, task exposure, time budget, and evaluation protocol aligned with the baseline, while adding only a lightweight self-evolving skill-library manager. This isolates the effect of reusable skills from that of changing the base model or supplying additional evidence. It is therefore a diagnostic condition rather than an optimized production agent, measuring whether agents can \textit{induce}, \textit{validate}, \textit{retrieve}, and \textit{reuse} compact analysis routines across the repeated release periods of a scenario.

\paragraph{Skill-library tools.} TimeSage-1.0 augments the baseline
\texttt{smolagent} harness with a skill-library layer, leaving the sandbox,
inputs, and required artifacts unchanged. It adds two tools: \texttt{induct\_skill}
saves a small reusable Python function with a name, description, and notes when
the agent identifies a recurring analysis idiom, and \texttt{retrieve\_skill}
returns the full body of a saved skill when its short description is insufficient for
reliable use.

\paragraph{Chronological registration.} Accepted skills become callable tools
 from the next period, so a skill induced in the period \(t_k\) cannot alter
the current answer but can affect behavior in \(t_{k+1}\) and beyond. This
preserves the chronological structure of the evolving environment.

\definecolor{tc0}{RGB}{204,220,210}
\definecolor{tc1}{RGB}{204,220,210}
\definecolor{tc2}{RGB}{204,220,210}
\definecolor{tc3}{RGB}{204,220,210}
\definecolor{tc4}{RGB}{204,220,210}
\definecolor{tc5}{RGB}{208,223,214}
\definecolor{tc6}{RGB}{211,225,216}
\definecolor{tc7}{RGB}{204,220,210}
\definecolor{tc8}{RGB}{237,242,238}
\definecolor{tc9}{RGB}{212,225,217}
\definecolor{tc10}{RGB}{204,220,210}
\definecolor{tc11}{RGB}{211,224,216}
\definecolor{tc12}{RGB}{204,220,210}
\definecolor{tc13}{RGB}{204,220,210}
\definecolor{tc14}{RGB}{204,220,210}
\definecolor{tc15}{RGB}{205,221,211}
\definecolor{tc16}{RGB}{204,220,210}
\definecolor{tc17}{RGB}{206,222,212}
\definecolor{tc18}{RGB}{204,220,210}
\definecolor{tc19}{RGB}{204,220,210}
\definecolor{tc20}{RGB}{205,220,211}
\definecolor{tc21}{RGB}{213,226,218}
\definecolor{tc22}{RGB}{215,227,219}
\definecolor{tc23}{RGB}{212,225,217}
\definecolor{tc24}{RGB}{211,225,216}
\definecolor{tc25}{RGB}{208,223,213}
\definecolor{tc26}{RGB}{204,220,210}
\definecolor{tc27}{RGB}{248,250,248}
\definecolor{tc28}{RGB}{206,222,212}
\definecolor{tc29}{RGB}{207,222,213}
\definecolor{tc30}{RGB}{206,221,212}
\definecolor{tc31}{RGB}{204,220,210}
\definecolor{tc32}{RGB}{216,228,220}
\definecolor{tc33}{RGB}{214,227,218}
\definecolor{tc34}{RGB}{210,224,215}
\definecolor{tc35}{RGB}{204,220,210}
\definecolor{tc36}{RGB}{223,233,226}
\definecolor{tc37}{RGB}{204,220,210}
\definecolor{tc38}{RGB}{216,228,220}
\definecolor{tc39}{RGB}{212,226,217}
\definecolor{tc40}{RGB}{213,226,218}
\definecolor{tc41}{RGB}{220,231,224}
\definecolor{tc42}{RGB}{233,240,235}
\definecolor{tc43}{RGB}{212,225,217}
\definecolor{tc44}{RGB}{219,230,223}
\definecolor{tc45}{RGB}{204,220,210}
\definecolor{tc46}{RGB}{215,227,219}
\definecolor{tc47}{RGB}{230,238,233}
\definecolor{tc48}{RGB}{209,224,214}
\definecolor{tc49}{RGB}{207,222,212}
\definecolor{tc50}{RGB}{214,227,218}
\definecolor{tc51}{RGB}{214,227,218}
\definecolor{tc52}{RGB}{228,236,231}
\definecolor{tc53}{RGB}{215,227,219}
\definecolor{tc54}{RGB}{217,229,221}
\definecolor{tc55}{RGB}{213,226,218}
\definecolor{tc56}{RGB}{221,232,225}
\definecolor{tc57}{RGB}{229,237,231}
\definecolor{tc58}{RGB}{215,227,219}
\definecolor{tc59}{RGB}{218,230,222}
\definecolor{tc60}{RGB}{233,240,235}
\definecolor{tc61}{RGB}{243,246,243}
\definecolor{tc62}{RGB}{244,247,245}
\definecolor{tc63}{RGB}{231,238,233}
\definecolor{tc64}{RGB}{237,243,239}
\definecolor{tc65}{RGB}{207,222,213}
\definecolor{tc66}{RGB}{211,225,216}
\definecolor{tc67}{RGB}{223,233,226}
\definecolor{tc68}{RGB}{204,220,210}
\definecolor{tc69}{RGB}{204,220,210}
\definecolor{tc70}{RGB}{227,236,230}
\definecolor{tc71}{RGB}{236,242,238}
\definecolor{tc72}{RGB}{243,247,244}
\definecolor{tc73}{RGB}{232,239,234}
\definecolor{tc74}{RGB}{234,240,236}
\definecolor{tc75}{RGB}{236,242,237}
\definecolor{tc76}{RGB}{237,243,239}
\definecolor{tc77}{RGB}{236,241,237}
\definecolor{tc78}{RGB}{231,239,234}
\definecolor{tc79}{RGB}{234,241,236}
\definecolor{tc80}{RGB}{248,250,248}
\definecolor{tc81}{RGB}{248,250,248}
\definecolor{tc82}{RGB}{233,240,235}
\definecolor{tc83}{RGB}{232,239,234}
\definecolor{tc84}{RGB}{240,245,241}
\definecolor{tc85}{RGB}{248,250,248}
\definecolor{tc86}{RGB}{248,250,248}
\definecolor{tc87}{RGB}{232,239,234}
\definecolor{tc88}{RGB}{235,241,237}
\definecolor{tc89}{RGB}{248,250,248}
\definecolor{tc90}{RGB}{248,250,248}
\definecolor{tc91}{RGB}{248,250,248}
\definecolor{tc92}{RGB}{248,250,248}
\definecolor{tc93}{RGB}{248,250,248}
\definecolor{tc94}{RGB}{248,250,248}
\definecolor{tc95}{RGB}{212,226,217}
\definecolor{tc96}{RGB}{222,233,226}
\definecolor{tc97}{RGB}{206,221,212}
\definecolor{tc98}{RGB}{214,227,219}
\definecolor{tc99}{RGB}{214,227,218}
\definecolor{tc100}{RGB}{244,247,245}
\definecolor{tc101}{RGB}{248,250,248}
\definecolor{tc102}{RGB}{248,250,248}
\definecolor{tc103}{RGB}{248,250,248}
\definecolor{tc104}{RGB}{248,250,248}
\definecolor{tc105}{RGB}{224,234,227}
\definecolor{tc106}{RGB}{207,222,212}
\definecolor{tc107}{RGB}{232,239,234}
\definecolor{tc108}{RGB}{248,250,248}
\definecolor{tc109}{RGB}{226,235,229}
\definecolor{tc110}{RGB}{228,237,231}
\definecolor{tc111}{RGB}{233,240,235}
\definecolor{tc112}{RGB}{237,242,238}
\definecolor{tc113}{RGB}{232,239,234}
\definecolor{tc114}{RGB}{232,239,234}
\definecolor{tc115}{RGB}{248,250,248}
\definecolor{tc116}{RGB}{248,250,248}
\definecolor{tc117}{RGB}{248,250,248}
\definecolor{tc118}{RGB}{248,250,248}
\definecolor{tc119}{RGB}{248,250,248}

\newcommand{\tval}[1]{#1}
\newcommand{\tvalb}[1]{\textbf{#1}}

\begin{table*}[t]
\centering
\caption{Per-difficulty-tier scores across evaluation dimensions. Cell shading uses a shared scale within each tier-and-metric group
(low~\scalebar~high); the best result in each group is shown in bold.}
\label{table_3}
\vspace{-5pt}
\setlength{\tabcolsep}{6pt}\small
\renewcommand{\arraystretch}{1.5}
\begin{tabularx}{\textwidth}{@{}ll>{\centering\arraybackslash}X>{\centering\arraybackslash}X>{\centering\arraybackslash}X>{\centering\arraybackslash}X>{\centering\arraybackslash}X@{}}
\toprule
\textbf{Model} & \textbf{Tier} & \textbf{Accuracy} & \textbf{Coverage} & \textbf{Faithfulness} & \textbf{Quality} & \textbf{Overall} \\
\midrule
\multirow{4}{*}{\logo{icon/openai.png}GPT-5.4} & Overall & \cellcolor{tc0}\tvalb{84.89} & \cellcolor{tc1}\tvalb{77.30} & \cellcolor{tc2}\tvalb{91.20} & \cellcolor{tc3}\tvalb{90.58} & \cellcolor{tc4}\tvalb{85.99} \\
 & Easy & \cellcolor{tc5}\tval{98.54} & \cellcolor{tc6}\tval{98.67} & \cellcolor{tc7}\tvalb{96.12} & \cellcolor{tc8}\tval{93.04} & \cellcolor{tc9}\tval{96.59} \\
 & Medium & \cellcolor{tc10}\tvalb{93.44} & \cellcolor{tc11}\tval{76.79} & \cellcolor{tc12}\tvalb{96.64} & \cellcolor{tc13}\tvalb{87.28} & \cellcolor{tc14}\tvalb{88.54} \\
 & Hard & \cellcolor{tc15}\tval{62.09} & \cellcolor{tc16}\tvalb{56.58} & \cellcolor{tc17}\tval{80.44} & \cellcolor{tc18}\tvalb{91.69} & \cellcolor{tc19}\tvalb{72.70} \\
\midrule
\multirow{4}{*}{\logo{icon/qwen.png}Qwen-3.5-397B} & Overall & \cellcolor{tc20}\tval{84.72} & \cellcolor{tc21}\tval{75.57} & \cellcolor{tc22}\tval{88.66} & \cellcolor{tc23}\tval{87.99} & \cellcolor{tc24}\tval{84.24} \\
 & Easy & \cellcolor{tc25}\tval{98.61} & \cellcolor{tc26}\tvalb{99.86} & \cellcolor{tc27}\tval{92.50} & \cellcolor{tc28}\tval{97.61} & \cellcolor{tc29}\tval{97.14} \\
 & Medium & \cellcolor{tc30}\tval{92.48} & \cellcolor{tc31}\tvalb{78.52} & \cellcolor{tc32}\tval{91.86} & \cellcolor{tc33}\tval{82.79} & \cellcolor{tc34}\tval{86.41} \\
 & Hard & \cellcolor{tc35}\tvalb{62.53} & \cellcolor{tc36}\tval{48.21} & \cellcolor{tc37}\tvalb{81.38} & \cellcolor{tc38}\tval{84.01} & \cellcolor{tc39}\tval{69.03} \\
\midrule
\multirow{4}{*}{\logo{icon/claude.png}Sonnet-4.6} & Overall & \cellcolor{tc40}\tval{82.76} & \cellcolor{tc41}\tval{74.32} & \cellcolor{tc42}\tval{84.46} & \cellcolor{tc43}\tval{87.97} & \cellcolor{tc44}\tval{82.38} \\
 & Easy & \cellcolor{tc45}\tvalb{99.47} & \cellcolor{tc46}\tval{97.99} & \cellcolor{tc47}\tval{93.95} & \cellcolor{tc48}\tval{97.19} & \cellcolor{tc49}\tval{97.15} \\
 & Medium & \cellcolor{tc50}\tval{89.03} & \cellcolor{tc51}\tval{75.91} & \cellcolor{tc52}\tval{87.21} & \cellcolor{tc53}\tval{82.43} & \cellcolor{tc54}\tval{83.65} \\
 & Hard & \cellcolor{tc55}\tval{59.36} & \cellcolor{tc56}\tval{49.03} & \cellcolor{tc57}\tval{72.04} & \cellcolor{tc58}\tval{84.75} & \cellcolor{tc59}\tval{66.29} \\
\midrule
\multirow{4}{*}{\logo{icon/kimi.png}Kimi-K2.6} & Overall & \cellcolor{tc60}\tval{77.88} & \cellcolor{tc61}\tval{70.07} & \cellcolor{tc62}\tval{81.74} & \cellcolor{tc63}\tval{81.90} & \cellcolor{tc64}\tval{77.90} \\
 & Easy & \cellcolor{tc65}\tval{98.75} & \cellcolor{tc66}\tval{98.65} & \cellcolor{tc67}\tval{94.59} & \cellcolor{tc68}\tvalb{97.97} & \cellcolor{tc69}\tvalb{97.49} \\
 & Medium & \cellcolor{tc70}\tval{82.89} & \cellcolor{tc71}\tval{69.88} & \cellcolor{tc72}\tval{81.22} & \cellcolor{tc73}\tval{74.40} & \cellcolor{tc74}\tval{77.10} \\
 & Hard & \cellcolor{tc75}\tval{51.68} & \cellcolor{tc76}\tval{41.82} & \cellcolor{tc77}\tval{69.52} & \cellcolor{tc78}\tval{74.00} & \cellcolor{tc79}\tval{59.25} \\
\midrule
\multirow{4}{*}{\logo{icon/mistral.png}Devstral-2-123B} & Overall & \cellcolor{tc80}\tval{74.14} & \cellcolor{tc81}\tval{69.08} & \cellcolor{tc82}\tval{84.43} & \cellcolor{tc83}\tval{81.37} & \cellcolor{tc84}\tval{77.26} \\
 & Easy & \cellcolor{tc85}\tval{89.70} & \cellcolor{tc86}\tval{92.28} & \cellcolor{tc87}\tval{93.82} & \cellcolor{tc88}\tval{93.30} & \cellcolor{tc89}\tval{92.27} \\
 & Medium & \cellcolor{tc90}\tval{73.11} & \cellcolor{tc91}\tval{66.81} & \cellcolor{tc92}\tval{79.34} & \cellcolor{tc93}\tval{67.03} & \cellcolor{tc94}\tval{71.57} \\
 & Hard & \cellcolor{tc95}\tval{59.76} & \cellcolor{tc96}\tval{48.44} & \cellcolor{tc97}\tval{80.56} & \cellcolor{tc98}\tval{84.96} & \cellcolor{tc99}\tval{68.43} \\
\midrule
\multirow{4}{*}{\logo{icon/gemma.png}Gemma-4-31B} & Overall & \cellcolor{tc100}\tval{75.05} & \cellcolor{tc101}\tval{69.08} & \cellcolor{tc102}\tval{80.87} & \cellcolor{tc103}\tval{76.25} & \cellcolor{tc104}\tval{75.31} \\
 & Easy & \cellcolor{tc105}\tval{95.03} & \cellcolor{tc106}\tval{99.42} & \cellcolor{tc107}\tval{93.82} & \cellcolor{tc108}\tval{91.37} & \cellcolor{tc109}\tval{94.91} \\
 & Medium & \cellcolor{tc110}\tval{82.23} & \cellcolor{tc111}\tval{70.76} & \cellcolor{tc112}\tval{83.79} & \cellcolor{tc113}\tval{74.48} & \cellcolor{tc114}\tval{77.81} \\
 & Hard & \cellcolor{tc115}\tval{47.43} & \cellcolor{tc116}\tval{37.07} & \cellcolor{tc117}\tval{64.82} & \cellcolor{tc118}\tval{63.09} & \cellcolor{tc119}\tval{53.10} \\
\bottomrule
\end{tabularx}
\end{table*}

\paragraph{Prompting and reuse.} The library is exposed through a period-level
skill-index block. When empty, it seeds the agent toward compact reusable
idioms, such as schema-specific loaders or row lookups, and away from one-off
arithmetic and literal answer dictionaries. Once non-empty, the block lists the
available skills, registers them as tools, and instructs the agent to invoke an
existing skill rather than re-implement it inline. The prompt templates are given in
the Appendix~\ref{appendix_prompts}.

\paragraph{Lightweight quality control.} Before a skill is exposed in later
periods, a check enforces that it is a callable, self-contained routine with a
usable interface and a faithful description, filtering out candidates that are
overly period-specific, duplicate an existing skill, or encode a literal answer.
The check is deliberately minimal, keeping the library usable during evaluation
without turning skill construction into a separate supervised process.

\subsection{Additional Results}
\label{appendix:additional_results}

\subsubsection{Breakdown of Main Results}
Table~\ref{table_3} provides a detailed breakdown of the performance of the model by difficulty tier, complementing the aggregate results reported in the main text. For each model, we report scores along four evaluation dimensions (Accuracy, Coverage, Faithfulness, and Quality) as well as the Overall score, separately for the Easy, Medium, and Hard tiers. Cell shading is computed within each tier-and-metric group, so colors are comparable across models within a column but not across tiers; the best entry in each group is shown in bold.

Two patterns are worth highlighting. First, performance degrades consistently from Easy to Hard across all models and dimensions: while most models exceed 90 Overall on the Easy tier, scores drop substantially on the Hard tier (from 72.70 for GPT-5.4 down to 53.02 for Gemma-4-31B), indicating that the harder questions remain a discriminating factor. Second, the dimension-level breakdown reveals that this degradation is driven primarily by Accuracy and Coverage, whereas Faithfulness and Quality remain comparatively stable. This suggests that on difficult inputs models tend to produce well-formed and faithful outputs that nonetheless miss or misidentify the correct content. GPT-5.4 and Qwen-3.5-397B are the strongest overall, while the gap between models widens on the Hard tier.

\subsubsection{LLM Judge Verification}
\label{appendix:llmjudge_verification}
To validate the reliability of our LLM judge, we conducted a human--judge agreement study on 300 instances (5 sampled periods per scenario) of GPT-5.4's agent outputs scored by DeepSeek-V4-Flash. Each instance was independently labeled by a human annotator---given the question, reference, and prediction---using the same discrete label space as the LLM judge: coverage (YES/PARTIAL/NO per keypoint), faithfulness (five fixed verdicts per atomic claim), and quality (hit/partial/miss/violation per signal). As shown in Table~\ref{tab:judge_agreement}, coverage and faithfulness exhibit strong alignment with human judgments, while quality agreement is lower yet still substantial. It is likely because quality assessment is inherently
more subjective, whereas the other two axes follow clearer, more objective rubrics. These results indicate that the LLM-judge setup can serve as a reliable proxy for human evaluation under our rubric.

\begin{table}[t]
\centering
\caption{Human--LLM judge agreement across the three evaluation axes.}
\label{tab:judge_agreement}
\begin{tabular}{lcc}
\toprule
\textbf{Evaluation Axis} & \textbf{Agreement Rate} & \textbf{Cohen's $\kappa$} \\
\midrule
Coverage     & 94.5\% & 0.882 \\
Faithfulness & 98.4\% & 0.917 \\
Quality      & 76.0\% & 0.665 \\
\bottomrule
\end{tabular}
\end{table}

\subsection{AI Usage}

We used AI assistance during the writing and editing process to improve the clarity, organization, and wording of the manuscript. The authors controlled scientific content, benchmark design, experimental analysis, and final writing decisions. AI tools were not used to generate benchmark ground truth without automatic and human verification. All reported results, figures, and tables are based on the authors' benchmark artifacts and evaluation pipeline.

\end{document}